\documentclass[10pt,twocolumn,letterpaper]{article}

\usepackage{style}
\date{}
\usepackage[pagebackref,breaklinks,colorlinks]{hyperref}

\usepackage{amsmath}
\usepackage{graphicx}
\usepackage{subcaption}
\usepackage{multirow}
\usepackage{arydshln}

\title{FERGI: Automatic Scoring of User Preferences for Text-to-Image Generation from Spontaneous Facial Expression Reaction}

\author{\parbox{16cm}{\centering
    {\large Shuangquan Feng\thanks{Equal Contribution}$^{\hspace{0.32em}1}$ \hspace{1cm} Junhua Ma\footnotemark[1]$^{\hspace{0.32em}2}$ \hspace{1cm} Virginia R. de Sa$^{3,4}$}\\
    {\normalsize
    $^1$ Neurosciences Graduate Program, University of California San Diego, La Jolla, USA\\
    $^2$ Department of Computer Science and Engineering, University of California San Diego, La Jolla, USA\\
    $^3$ Department of Cognitive Science, University of California San Diego, La Jolla, USA\\
    $^4$ Halıcıoğlu Data Science Institute, University of California San Diego, La Jolla, USA}}
}

\usepackage{placeins}
\begin{document}
\maketitle
\begin{abstract}
Researchers have proposed to use data of human preference feedback to fine-tune text-to-image generative models. However, the scalability of human feedback collection has been limited by its reliance on manual annotation. Therefore, we develop and test a method to automatically score user preferences from their spontaneous facial expression reaction to the generated images. We collect a dataset of Facial Expression Reaction to Generated Images (FERGI) and show that the activations of multiple facial action units (AUs) are highly correlated with user evaluations of the generated images. We develop an FAU-Net (Facial Action Units Neural Network), which receives inputs from an AU estimation model, to automatically score user preferences for text-to-image generation based on their facial expression reactions, which is complementary to the pre-trained scoring models based on the input text prompts and generated images. Integrating our FAU-Net valence score with the pre-trained scoring models improves their consistency with human preferences. This method of automatic annotation with facial expression analysis can be potentially generalized to other generation tasks. The code is available at https://github.com/ShuangquanFeng/FERGI, and the dataset is also available at the same link for research purposes.
\end{abstract}
\section{Introduction}
The rapid recent advancements in text-to-image generative models have enabled the generation of high-fidelity images aligned with text prompts \cite{nichol2021glide,ramesh2021zero,saharia2022photorealistic,ramesh2022hierarchical,rombach2022high, podell2023sdxl}. To better improve the fidelity of the generated images and their alignment with the text prompts, researchers collected large-scale datasets of human preference feedback on images generated by text-to-image models \cite{xu2023imagereward,li2023agiqa,kirstain2023pick,wu2023human,wu2023human2,zhang2023perceptual} and proposed methods of fine-tuning the models with data of human preference feedback \cite{xu2023imagereward,lee2023aligning,fan2023dpok}.

However, the scalability of human feedback collection has still been limited by its reliance on manual annotation. Researchers have used the CLIP score \cite{radford2021learning}, Aesthetic score \cite{schuhmann2022laion}, BLIP score \cite{li2022blip}, ImageReward score \cite{xu2023imagereward}, PickScore \cite{kirstain2023pick}, and HPS v2 score \cite{wu2023human2} to automatically score human preferences of text-to-image generation based on the input text prompts and the generated images. In this study, we develop and test a system to automatically score user preferences from their spontaneous facial expression reaction to the generated images, which is complementary to the models based only on the texts and images. This automatic scoring requires zero additional effort from the real users of text-to-image generative models.

We present the Facial Expression Reaction to Generated Images (FERGI) dataset, which comprises  video recordings of 33 participants' facial expression reaction to 2827 images generated by Stable Diffusion (SD) v1.4 \cite{rombach2022high} based on 576 different self-drafted text prompts, along with their feedback on the generated images from manual input. We estimate the activation of facial action units (AUs), as defined in the Facial Action Coding System (FACS, a comprehensive system breaking down facial expressions into individual components of AUs of muscle movements \cite{ekman1978facial}), in the facial expression reaction videos with a model we trained on external datasets and show that the activations of multiple AUs are highly correlated with user evaluations of the generated images. We further propose a method of automatically scoring user preferences by detecting the activations of the AUs in the facial expression reaction to the generated images and feeding them into a neural network.
\section{Related Work}
\textbf{Text-to-Image Generation and Evaluation. }Various models have been developed for text-to-image generation, including Generative Adversarial Networks (GANs) \cite{goodfellow2020generative,karras2020analyzing,brock2018large,karras2019style,ding2021cogview,ding2022cogview2}, Variational autoencoders (VAEs) \cite{kingma2013auto,child2020very,vahdat2020nvae}, flow-based models \cite{dinh2014nice,dinh2016density,kingma2018glow}, autoregressive models (ARMs) \cite{chen2020generative,child2019generating, van2016conditional,van2016pixel,yu2022scaling}, and diffusion models (DMs) \cite{sohl2015deep,ho2020denoising,song2019generative}. More recent advancements in DMs \cite{nichol2021glide,rombach2022high,ramesh2022hierarchical,podell2023sdxl,saharia2022photorealistic,dhariwal2021diffusion} have achieved great success in generating high-quality images and attracted widespread public attention. As traditional evaluation metrics like Inception Score (IS) \cite{barratt2018note}, Fréchet Inception Distance (FID) \cite{heusel2017gans}, and CLIP score \cite{radford2021learning} failed to comprehensively capture human preferences for text-to-image generation, researchers have proposed to specifically train human preference scoring models based on large-scale human feedback datasets, such as Human Preference Score (HPS) \cite{wu2023human}, ImageReward \cite{xu2023imagereward}, PickScore \cite{kirstain2023pick}, and Human Preference Score v2 (HPS v2) \cite{wu2023human2}, and demonstrated their effectiveness to improve text-to-image generative models \cite{xu2023imagereward,kirstain2023pick,wu2023human2,fan2023dpok}.

\textbf{Human Feedback Datasets. } There are multiple datasets of human feedback on text-to-image generation, including AGIQA-1K \cite{zhang2023perceptual}, Human Preference Dataset (HPD) \cite{wu2023human}, ImageReward \cite{xu2023imagereward}, Pick-a-Pic \cite{kirstain2023pick}, AGIQA-3K \cite{li2023agiqa}, and Human Preference Dataset v2 (HPD v2) \cite{wu2023human2}. To the best of our knowledge, our FERGI dataset is the first to include both manual annotations of generated images and the associated facial expression reaction videos.

\textbf{Facial Expression Recognition and Applications.}
Automatic facial expression recognition has advanced rapidly in recent years \cite{li2020deep}. As emotional responses to image generation can be more complicated than single categories of the most widely-researched seven basic emotions \cite{ekman1971constants,ekman1994strong,matsumoto1992more}, we decided to analyze the facial expression reactions directly in terms of muscle movements, as defined by AUs in FACS \cite{ekman1978facial}. AU detection and estimation has attracted increasing interest \cite{jacob2021facial,shao2019facial,zhao2016deep,zhao2015joint,chu2017learning,li2017action,li2018eac,walecki2017deep,kollias2022abaw,kollias2023abaw,zhi2020comprehensive,martinez2017automatic,valstar2015fera,valstar2017fera} and been used for facial emotion recognition \cite{xu2018chapter,yang2019facial,XuNIPS2019,XuFG2020,XudeSa21}. For example, researchers have shown that the activation of the corrugator (activated in AU4) and zygomaticus (activated in AU12) are associated with negative and positive emotions respectively and are the most investigated muscle activations in studying affect and emotion \cite{FacialEMG}. Also, the activity of the corrugator supercilii (activated in AU4) is positively associated with amygdala and negatively associated with ventromedial prefrontal cortex activity \cite{heller2014face} and increases with negative stimuli, such as negative images \cite{cacioppo1986electromyographic,cacioppo1990inferring,larsen2003effects,lang1993looking}. For applications, \cite{becattini2021plm,wendin2011facial} showed the effectiveness of using facial expression to infer/analyze human preferences.
\section{FERGI Dataset}
\label{sec:FERGI}

\subsection{Participants}
39 participants were recruited from the SONA system of University of California San Diego (UCSD) and completed the study asynchronously on personal computers. 6 participants were removed from the dataset for various reasons, including failure to participate, failure in video recording, and not allowing sharing of their video for research. Data collection and use were approved by the Institutional Review Board (IRB) of UCSD.

\subsection{Data Collection Procedure}
\begin{figure*}[tb]
  \centering
  \includegraphics[width=.95\linewidth]{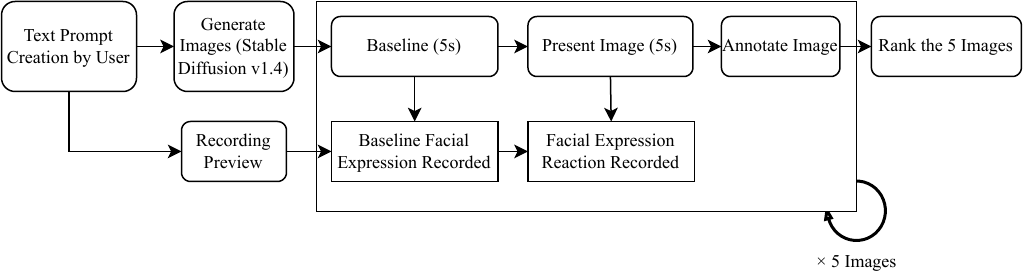}
  \caption{\textbf{Procedure of one session in data collection.} Firstly, the participant creates an input text prompt. Secondly, the participant is directed to a webcam preview to confirm that the webcam captures their face appropriately; at the same time, 5 images are generated from the text prompt using Stable Diffusion v1.4. Then, for each image, the participant goes through a 5-second baseline phase, a 5-second image presentation phase, and an image annotation phase with no time restriction. (During the baseline phase, the participant's baseline facial expression is recorded, and during the image presentation phase, the participant's facial expression reaction to the generated image is recorded.) Finally, after all 5 images have been presented and annotated, the participant ranks the 5 generated images from best to worst. A copy of this figure with additional references to the screenshots of each stage is in the supplemental material.}
  \label{fig:data_collection_procedure}
\end{figure*}

Each participant completed multiple sessions of data collection. In each session, the participant drafted one text prompt and viewed 5 images generated from the input text prompt. A flow chart of the procedure of each session is shown in \Cref{fig:data_collection_procedure}. Details are explained below.

\subsubsection{Prompt Creation}
There are two types of sessions, structured input and free-form input. In both types, the participant can freely draft any text prompt as long as it is not NSFW. The only difference lies in how the input text prompt is created. In a structured input session, the participant fills out a form to separately specify different elements desired in the generated images: animate objects, inanimate objects, interactional relations (between two objects), positional relations (between two objects), location, style (of the image), and keywords. Based on the contents of the form, an initial text prompt is automatically generated by the Large Language Model (LLM) OpenAI GPT-3.5 Turbo which the participant can freely adjust to obtain the final input text prompt. In a free-form input, the participant directly enters the entire final input text prompt.

\subsubsection{Recording}

After the input text prompt is finalized, the participant is directed to a webcam preview to confirm that the webcam captures their face appropriately. The webcam is used to record throughout the rest of the session. The configuration of the raw videos varied based on the participant’s device and browser. However, recorded videos of most participants have a resolution of $640\times 480$, and all recorded videos were standardized to 30 FPS afterwards.

\subsubsection{Image Presentation, Annotation, and Ranking}
In each session, a total of 5 images generated by SD v1.4 \cite{rombach2022high} from the input text prompt are presented and annotated sequentially and ranked afterwards (present image 1 $\rightarrow$ annotate image 1 $\rightarrow$ $\dots$ $\rightarrow$ present image 5 $\rightarrow$ annotate image 5 $\rightarrow$ rank 5 images). The image presentation and annotation followed the webcam preview. The details are as follows:
\begin{itemize}
\item \textbf{Baseline} (5 seconds): The website displays the ``preparing image'' status text at the center of the page along with the input text prompt. No specific facial-expression-eliciting event is expected during this period. 
\item \textbf{Present} (5 seconds): The generated image is presented at the center of the webpage along with the input text prompt. The participant's facial expression reaction to the generated image is collected within this interval.
\item \textbf{Annotate} (no time restriction): The website displays the prompt, the generated image (moved to the right side), and the annotation survey. The participant annotates the image manually by filling out the survey during this interval. The survey was adapted from the survey designed for ImageReward \cite{xu2023imagereward} and includes the following questions: important elements not reflected in the image (free-response question), overall rating (on a scale of 1 (worst) to 7 (best)), image-text alignment rating (on a scale of 1 to 7), fidelity rating (on a scale of 1 to 7), issues of the image\footnote{For the first 7 out of 33 participants (or 8 out of the total 39 participants), there was a bug (later fixed) in checkbox selection for answering the questions regarding ``issues of the image'' that made the recorded answers unreliable, so answers for this question should not be used in analysis.\label{fn:bug_in_issues}} (multi-answer question), and emotions felt when seeing the image (multi-answer question). 
\end{itemize}
After the annotation of the last image, the participant ranks the 5 generated images from best to worst, and the data of the session is uploaded to the web server ending the session. 

\subsection{Data Preprocessing}
For each image, two clips are extracted from the recorded video: a 5-second ``baseline clip'' of the participant's facial expression during the baseline period and a 5-second ``reaction clip'' of the participant's facial expression during the image presentation period.

SD v1.4 \cite{rombach2022high} outputs an all-black image when it detects the original output image might be inappropriate. A total of 53 such images out of 2880 total generated were generated in the dataset. The data associated with these images were excluded from further analysis, resulting in 
fewer than 5 valid images for some input text prompts.
\section{AU Model Training}
\label{sec:ModelTraining}
We analyze the facial expression reaction of the users by estimating their AU activations. We trained our AU estimation model on the DISFA  \cite{mavadati2013disfa,mavadati2012automatic} and DISFA+ \cite{mavadati2016extended} datasets. They are among the largest, most popular, and most cited AU datasets with intensity annotations. In these datasets, each frame is manually annotated by a human expert with intensities on a scale of 0 to 5, for activation of AU1 (inner brow raiser), AU2 (outer brow raiser), AU4 (brow lowerer), AU5 (upper lid raiser), AU6 (cheek raiser), AU9 (nose wrinkler), AU12 (lip corner puller), AU15 (lip corner depressor), AU17 (chin raiser), AU20 (lip stretcher), AU25 (lips part), and AU26 (jaw drop). (See the supplemental material for a visual reference guide for the analyzed AUs.) Video frames are first preprocessed with face detection \cite{lugaresi2019mediapipe}, face alignment \cite{PFL}, and a combination of histogram equalization and linear mapping \cite{kuo2018compact} before being fed into the model for training, testing, or inference. For model training, we used the neural network IR-50 \cite{deng2019arcface} pre-trained on Glink360k \cite{an2022killing} and fine-tuned it on the DISFA and DISFA+ datasets \cite{mavadati2013disfa,mavadati2016extended}. A single network learned the estimation of all labeled AUs  with both regression and ordinal classification \cite{niu2016ordinal}. More details of the training of the AU model are elaborated in the supplemental material.

The comparison between our AU estimation model's performance on the DISFA dataset and other state-of-the-art models is shown in \Cref{tab:AU_model_performances}. Our model's ICC(3,1) (higher is better) is the best among all models. Although our model's MAE (lower is better) is less competitive among all models, we believe that ICC(3,1) is a better metric for evaluating the performance of AU estimation models on the DISFA dataset considering the high imbalance of the dataset. Also, ICC(3,1) is specifically more important in our application here because we care about the change of facial expression in reaction to the generated images instead of its absolute intensity. 

\begin{table*}[tb]
    \caption{Comparison of our AU estimation model's performance on the DISFA dataset with other models}
    \centering
    \begin{tabular}{|c|cccccccccccccc|}
        \hline
        \multirow{2}{*}{Metric} & \multirow{2}{*}{Method} & \multicolumn{12}{c}{AU} & \multirow{2}{*}{Average}\\
        \cline{3-14}
        & & 1 & 2 & 4 & 5 & 6 & 9 & 12 & 15 & 17 & 20 & 25 & 26 & \\ \hline
        \multirow{5}{*}{ICC(3,1)$\uparrow$} & 
        CCNN-IT \cite{walecki2017deep} & .18 & .15 & .61 & .07 & \textbf{.65} & .55 & .82 & .44 & .37 & \textbf{.28} & .77 & .54 & .45\\
        & 2DC \cite{linh2017deepcoder} & .70 & \textbf{.55} & .69 & .05 & .59 & \textbf{.57} & \textbf{.88} & .32 & .10 & .08 & .90 & .50 & .50\\
        & SCC-Heatmap \cite{fan2020facial} & \textbf{.73} & .44 & .74 & .06 & .27 & .51 & .71 & .04 & .37 & .04 & .94 & \textbf{.78} & .47\\
        & iARL \cite{shao2019facial} & .13 & .36 & .68 & .22 & .56 & .36 & .86 & \textbf{.52} & .37 & .12 & \textbf{.96} & .60 & .48\\
        \cdashline{2-15}
        & Our Model & .53 & .45 & \textbf{.75} & \textbf{.62} & .55 & \textbf{.57} & .84 & .42 & \textbf{.47} & .24 & .93 & .65 & \textbf{.59}\\
        \hline
        
        \multirow{4}{*}{MAE$\downarrow$} & 
        CCNN-IT \cite{walecki2017deep} & 87 & .63 & .86 & .26 & .73 & .57 & .55 & .38 & .57 & .45 & .81 & .64 & .61\\
        & SCC-Heatmap \cite{fan2020facial} & \textbf{.16} & \textbf{.16} & \textbf{.27} & \textbf{.03} & \textbf{.25} & \textbf{.13} & .32 & .15 & \textbf{.20} & .09 & .30 & .32 & \textbf{.20}\\
        & iARL \cite{shao2019facial} & .30 & .31 & .52 & .04 & .36 & .30 & \textbf{.31} & \textbf{.05} & .33 & \textbf{.08} & \textbf{.29} & \textbf{.26} & .26\\
        \cdashline{2-15}
        & Our Model & .37 & .39 & .44 & .11 & .35 & .21 & .34 & .20 & .39 & .21 & .32 & .42 & .31\\
        \hline
    \end{tabular}
    \label{tab:AU_model_performances}
\end{table*}
\section{Facial Feature Extraction}

In this section, we elaborate the procedure of extracting facial features from the video clips associated with each generated image based on the trained AU recognition model. Similar procedures were used to extract features from baseline clips and reaction clips. Further details are in the supplemental material.

\subsection{Data Filtering}
\label{sec:DataFiltering}
Robustness against face occlusion and faces with non-frontal poses is an ongoing challenge for facial expression recognition models \cite{zhang2018facial,jampour2022multiview}. Our AU model was trained on datasets consisting predominantly of frontal facial images without occlusion and thus for each 5-second video clip, we exclude frames with a low face detection confidence score (given by MediaPipe \cite{lugaresi2019mediapipe}) and frames with an off-angle pitch or yaw (estimated based on detected facial landmarks \cite{PFL}). If a clip has more than $20\%$ of its total frames excluded, the whole clip, and its associated data (survey and ranking results) are excluded from further analysis. 

\subsection{AU Activation Value}
We compute an AU activation value $\alpha_{i}$ for each trained AUi in each video clip. The AU activation value is computed as follows:
\begin{itemize}
    \item Firstly, the intensity of each AU for each frame of the video clip is estimated with the AU model: $\hat{y}_{i,\mathrm{reg}}^{(1)}, \hat{y}_{i,\mathrm{reg}}^{(2)}, \dots$, where the superscript indicates the index of the frame.
    \item Secondly, a moving window mean of the estimated AU intensities for every 0.1 seconds (3 frames for 30 FPS) is computed:
    \begin{equation}
        \bar{\hat{y}}_{i,\mathrm{reg}}^{(k)}=\frac{1}{3}\sum_{k'=k}^{k+2}\hat{y}_{i,\mathrm{reg}}^{(k')}
        \label{eqn:AU_moving_window_mean}
    \end{equation}
    for all $k$ with defined $\hat{y}_{i,\mathrm{reg}}^{(k+1)}$ and $\hat{y}_{i,\mathrm{reg}}^{(k+2)}$ (as they may be undefined for frames at the end of the video clip or frames adjacent to excluded frames).
    \item Finally, the AU activation of the clip as the difference between the maximum AU intensity within the video clip and the AU intensity of the first 0.1 seconds is computed:
    \begin{equation}
        \alpha_{i} = \max_{k}(\bar{\hat{y}}_{i,\mathrm{reg}}^{(k)}) - \bar{\hat{y}}_{i,\mathrm{reg}}^{(1)}.
    \end{equation}
\end{itemize}
\section{Experiments}
\subsection{Statistical Analysis}
\begin{figure*}[tb]
  \centering
  \begin{subfigure}{0.24\linewidth}
    \includegraphics[width=\linewidth]{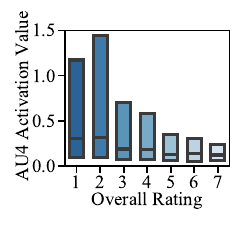}
    \caption{AU4}
\label{fig:overall_rating_and_AU4_activation_value}
  \end{subfigure}
  \centering
  \begin{subfigure}{0.24\linewidth}
    \includegraphics[width=\linewidth]{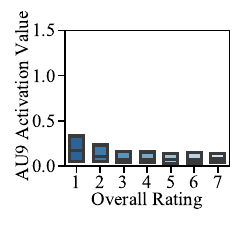}
    \caption{AU9}
\label{fig:overall_rating_and_AU9_activation_value}
  \end{subfigure}
  \begin{subfigure}{0.24\linewidth}
    \includegraphics[width=\linewidth]{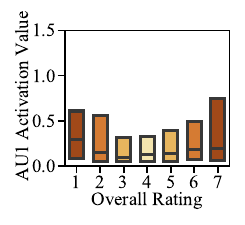}
    \caption{AU1}
\label{fig:overall_rating_and_AU1_activation_value}
  \end{subfigure}
    \begin{subfigure}{0.24\linewidth}
    \includegraphics[width=\linewidth]{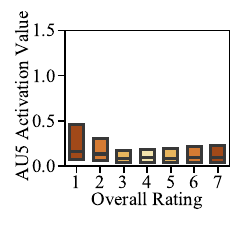}
    \caption{AU5}
\label{fig:overall_rating_and_AU5_activation_value}
  \end{subfigure}
  \centering
    \begin{subfigure}{0.24\linewidth}
    \includegraphics[width=\linewidth]{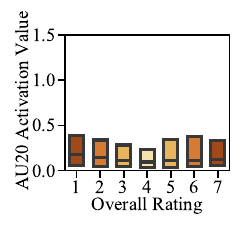}
    \caption{AU20}
\label{fig:overall_rating_and_AU20_activation_value}
  \end{subfigure}
  \begin{subfigure}{0.24\linewidth}
    \includegraphics[width=\linewidth]{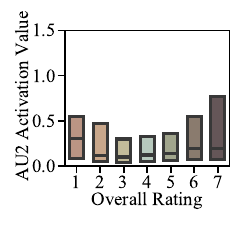}
    \caption{AU2}
\label{fig:overall_rating_and_AU2_activation_value}
  \end{subfigure}
 \centering
  \begin{subfigure}{0.24\linewidth}
    \includegraphics[width=\linewidth]{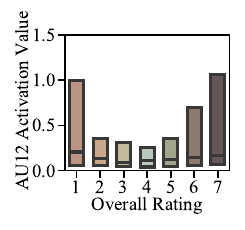}
    \caption{AU12}
\label{fig:overall_rating_and_AU12_activation_value}
  \end{subfigure}
  \caption{\textbf{Overall ratings highly correlated with AU activation values.} The distributions of the activation values of multiple AUs for images of different ratings. Each subfigure shows the results for a different AU (indicated in the captions) and contains 7 boxplots for AU activation values (on the y-axis) for images with 7 different ratings along the x-axis. In each boxplot, the bottom/top of the box represents the first/third quartile (25th/75th percentile) of the AU activation values, and the line in the middle of the box represents the median. Blue bars are used for AUs that are significantly correlated with the ratings while brown bars are used for AUs that are significantly correlated with the extremity of the ratings (computed as $\left|\text{overall rating} - 4\right|$).  AU2 and AU12 are shown in blue and brown reflecting significance (after multiple comparisons) for both ratings and extremity of ratings.  The darker blue and darker brown colors show the ratings associated with higher AU activations.
  The numbers of ratings from 1 to 7 are 104, 185, 258, 456, 581, 403, and 216 respectively.} \label{fig:overall_rating_and_AU_activation_values}
\end{figure*}

\begin{figure}[htbp]
  \centering
  \begin{subfigure}{0.49\linewidth}
    \includegraphics[scale=1]{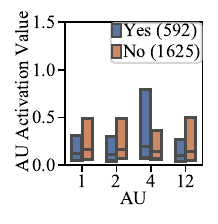}
    \caption{disappointed}
    \label{fig:disappointed_and_AU_activation_value}
  \end{subfigure}
  \centering
  \begin{subfigure}{0.49\linewidth}
    \includegraphics[scale=1]{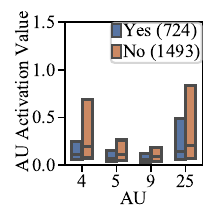}
    \caption{satisfied}
    \label{fig:satisfied_and_AU_activation_value}
  \end{subfigure}\\
  \begin{subfigure}{0.49\linewidth}
    \includegraphics[scale=1]{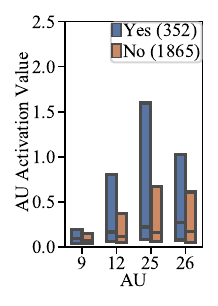}
    \caption{surprised}
    \label{fig:surprised_and_AU_activation_value}
  \end{subfigure}
  \centering
  \begin{subfigure}{0.49\linewidth}
    \includegraphics[scale=1]{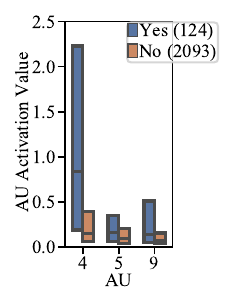}
    \caption{disgusted}
    \label{fig:disgusted_and_AU_activation_value}
  \end{subfigure}\\
  \centering
  \begin{subfigure}[t]{0.51\linewidth}
    \includegraphics[scale=1]{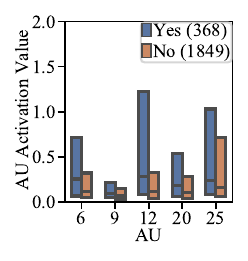}
    \caption{amused}
    \label{fig:amused_and_AU_activation_value}
  \end{subfigure}
  \centering
  \begin{subfigure}[t]{0.47\linewidth}
    \includegraphics[scale=1]
    {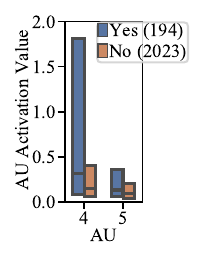}
    \caption{scared}
    \label{fig:scared_and_AU_activation_value}
  \end{subfigure}
  \caption{\textbf{Reported emotions highly correlated with AU activation values.} The distributions of the activation values of multiple AUs for images eliciting different emotions of the participants as reported in  answers to the question ``Did you feel any of the following emotions when you saw the image?'' Each subfigure shows the results for a different emotion (indicated in the subcaptions) and contains multiple pairs of boxplots representing the results for different AUs. The x-axis represents the indices of the AUs while the y-axis represents the AU activation values in the reaction clip of each image. In each pair of boxplots, the boxplot on the left/right side represents the AU activation values for images that did/didn't elicit the corresponding emotion. In each boxplot, the bottom of the box represents the first quartile (25th percentile) of the AU activation values, the top of the box represents the third quartile (75th percentile), and the line in the middle of the box represents the median. 
  The numbers of responses of each type for each subfigure are as follows. 
  (a) 
  Yes (592); No (1625). 
  (b) 
  Yes (724); No (1493). 
  (c) 
  Yes (352); No (1865). 
  (d) 
  Yes (124); No (2093). 
  (e) 
  Yes (368); No (1849). 
  (f)  
  Yes (194); No (2023).
  } 
  \label{fig:emotions_and_AU_activation_values}
\end{figure}

We started with analyzing statistical relationships between the AU activation values and answers in the annotation survey for each image. The primary results are as follows:
\begin{itemize}
    \item \textbf{Overall Ratings.} The participant's overall ratings (on a scale of 1 to 7) of the generated images have a significant positive correlation with the activation values of AU2 and AU12 and a significant negative correlation with the activation values of AU4 and AU9 as shown in \Cref{fig:overall_rating_and_AU2_activation_value,fig:overall_rating_and_AU4_activation_value,fig:overall_rating_and_AU9_activation_value,fig:overall_rating_and_AU12_activation_value} (Spearman correlation: $\rho\approx 0.09$ and $p<5\times 10^{-5}$ for AU2,
    $\rho\approx -0.18$ and $p<5\times 10^{-17}$ for AU4,
    $\rho\approx -0.09$ and $p<5\times 10^{-5}$ for AU9,
    $\rho\approx 0.08$ and $p<5\times 10^{-4}$ for AU12).
    \item \textbf{Extremity of Overall Ratings.} The 
    magnitude of deviation from the midpoint rating ($\left|\text{overall rating} - 4\right|$) have a significant positive correlation with the activation values of AU1, AU2, AU5, AU12, and AU20 as shown in \Cref{fig:overall_rating_and_AU1_activation_value,fig:overall_rating_and_AU2_activation_value,fig:overall_rating_and_AU5_activation_value,fig:overall_rating_and_AU12_activation_value,fig:overall_rating_and_AU20_activation_value} (Spearman correlation: $\rho\approx 0.10$ and $p<5\times 10^{-6}$ for AU1, $\rho\approx 0.11$ and $p<5\times 10^{-7}$ for AU2, $\rho\approx 0.08$ and $p<5\times 10^{-4}$ for AU5, $\rho\approx 0.14$ and $p<5\times 10^{-11}$ for AU12, $\rho\approx 0.08$ and $p<1\times 10^{-4}$ for AU20). \item \textbf{Emotions.} For the question ``Did you feel any of the following emotions when you saw the image?'', the participant can independently select 0 to 6 answers from the six options: surprised, disgusted, amused, scared, disappointed, and satisfied. Each of the reported emotions was found to be significantly correlated with the activation values of multiple AUs.
    \begin{itemize}
        \item \textbf{Disappointed.} ``Disappointing'' images are associated with lower activation values of AU1, AU2, and AU12 and higher activation values of AU4 as shown in \Cref{fig:disappointed_and_AU_activation_value} (Wilcoxon rank-sum tests: $z\approx -4.25$ and $p<5\times 10^{-5}$ for AU1, $z\approx -5.82$ and $p<1\times 10^{-8}$ for AU2, $z\approx 4.82$ and $p<5\times 10^{-6}$ for AU4, $z\approx -4.86$ and $p<5\times 10^{-6}$ for AU12).
        \item \textbf{Satisfied.} ``Satisfying'' images are associated with lower activation values of AU4, AU5, AU9, and AU25 as shown in \Cref{fig:satisfied_and_AU_activation_value} (Wilcoxon rank-sum tests: $z\approx -9.07$ and $p<5\times 10^{-19}$ for AU4, $z\approx -6.62$ and $p<5\times 10^{-11}$ for AU5, $z\approx -6.93$ and $p<5\times 10^{-12}$ for AU9, $z\approx -3.79$ and $p<5\times 10^{-4}$ for AU25).
        \item \textbf{Surprised.} ``Surprising'' images are associated with higher activation values of AU9, AU12, AU25, and AU26 as shown in \Cref{fig:surprised_and_AU_activation_value} (Wilcoxon rank-sum tests: $z\approx 3.78$ and $p<5\times 10^{-4}$ for AU9, $z\approx 4.29$ and $p<5\times 10^{-5}$ for AU12, $z\approx 4.01$ and $p<1\times 10^{-4}$ for AU25, $z\approx 4.45$ and $p<1\times 10^{-5}$ for AU26).
        \item \textbf{Disgusted.} ``Disgusting'' images are associated with higher activation values of AU4, AU5, and AU9 as shown in \Cref{fig:disgusted_and_AU_activation_value} (Wilcoxon rank-sum tests: $z\approx 8.43$ and $p<1\times 10^{-16}$ for AU4, $z\approx 3.94$ and $p<1\times 10^{-4}$ for AU5, $z\approx 4.87$ and $p<1\times 5^{-6}$ for AU9).
        \item \textbf{Amused.} ``Amusing'' images are associated with higher activation values of AU6, AU9, AU12, AU20, and AU25 as shown in \Cref{fig:amused_and_AU_activation_value} (Wilcoxon rank-sum tests: $z\approx 7.00$ and $p<1\times 10^{-11}$ for AU6, $z\approx 5.01$ and $p<1\times 10^{-6}$ for AU9, $z\approx 9.15$ and $p<1\times 10^{-19}$ for AU12, $z\approx 6.12$ and $p<1\times 10^{-9}$ for AU20, $z\approx 3.85$ and $p<1\times 10^{-4}$ for AU25).
        \item \textbf{Scared.} ``Scary'' images are associated with higher activation values of AU4 and AU5 as shown in \Cref{fig:scared_and_AU_activation_value} (Wilcoxon rank-sum tests: $z\approx 5.60$ and $p<5\times 10^{-8}$ for AU4, $z\approx 4.39$ and $p<5\times 10^{-5}$ for AU5).
    \end{itemize}
\end{itemize}
Note that the results presented above selectively include the AUs whose associated p-values (as reported before corrections) are below the Bonferroni-corrected significance threshold of $0.05/[12\times(2+6)]$ (12 investigated AUs, 2 tests for overall ratings (1 for the raw rating and 1 for the extremity of the rating), and 6 emotions).

\subsection{Automatic Annotation of User Preferences between Image Pairs}
\label{sec:Automatic_Annotation_of_User_Preference_between Image_Pairs_Using_FAU-Net}

\subsubsection{Pre-trained Scoring Models}
\label{sec:pretrained_scoring_models}
Researchers have used the CLIP score \cite{radford2021learning}, Aesthetic score \cite{schuhmann2022laion}, BLIP score \cite{li2022blip}, ImageReward score \cite{xu2023imagereward}, PickScore \cite{kirstain2023pick}, and HPS v2 score \cite{wu2023human2} to estimate human preferences of text-to-image generation based on input text prompts and generated images.

Among these baseline scoring models, ImageReward, PickScore, and HPS v2 were all trained specifically on datasets of human preferences of text-to-image generation and are supposed to outperform the other models \cite{xu2023imagereward,kirstain2023pick,wu2023human2}. To introduce a potentially more competitive baseline score, we further combine them to form an ensemble baseline score
\begin{align}
    s_\mathrm{ens} = w_\mathrm{IR}s_\mathrm{IR} + w_\mathrm{Pick}s_\mathrm{Pick} + w_\mathrm{HPSv2}s_\mathrm{HPSv2} & \label{eqn:define_ensemble_baseline_score} \\
    \text{subject to} \quad w_\mathrm{IR} + w_\mathrm{Pick} + w_\mathrm{HPSv2} & = 1,
    \label{eqn:define_ensemble_baseline_score_constrain}
\end{align}
where $s_\mathrm{ens}$ represents the ensemble baseline score, $s_\mathrm{IR}$, $s_\mathrm{Pick}$, and $s_\mathrm{HPSv2}$ represent the ImageReward score, PickScore, and HPS v2 score respectively, and $w_\mathrm{IR}$, $w_\mathrm{Pick}$, and $w_\mathrm{HPSv2}$ represent their respective weights.

A leave-one-participant-out (LOPO) procedure was applied in the fitting of the three weights $w_\mathrm{IR}$, $w_\mathrm{Pick}$, and $w_\mathrm{HPSv2}$; we used grid search with a granularity of 0.1 to optimize the accuracy of image preference prediction. Since the scales of the scores are different, they were each z-scored using their mean and standard deviation on the training set. We obtained the same results of $w_\mathrm{IR}=0.1$, $w_\mathrm{Pick}=0.6$, and $w_\mathrm{HPSv2}=0.3$ for all LOPO training sets.

\subsubsection{FAU-Net Valence Score}
We aim to use the estimated AU activation values to automatically score user preferences between images generated from the same input text prompt and propose that this scoring more directly reflects the user's reaction to the generated image and is complementary to the pre-trained scoring models. To achieve this goal, we train an FAU-Net (Facial Action Units Neural Network) to do automatic scoring. In the FAU-Net, the estimated activation values of the 12 investigated AUs are used as the inputs; each activation value is first preprocessed with a linear transformation with trained parameters and sigmoid function ($\sigma(a\cdot \alpha+b)$ with the initializations of $a=1$ and $b=-1$ for encouraging transformations of positive correlation for better interpretability); then the preprocessed AU activation values are used as inputs to a hidden layer with 16 neurons which feeds into a single output neuron representing the AU evaluation score of the image; this output unit receives inputs from both the 16 neurons of the hidden layer and the 12 preprocessed AU activation values. The sigmoid function is used as the activation function for all neurons. The network is trained with a RankNet loss function \cite{burges2005learning} using the reaction clips and ranking order given by the participant for image pairs generated from the same text prompts and an L1 regularization with a coefficient of $10^{-4}$. For model training, we use the Adam optimizer with an initial learning rate of $10^{-3}$ and a single batch for all data to train the neural network for 300 epochs using a single NVIDIA RTX A6000 48GB GDDR6 GPU for about 15 min.

Our FAU-Net is trained with an LOPO procedure and outputs an FAU-Net valence score $s_\mathrm{FAU}$ reflecting the evaluation of the generated image derived from the facial expression reaction. Note that the AU estimation model is frozen and not trained with the FERGI data.

\subsubsection{Integrate FAU-Net Valence Score with Pre-trained Scoring Models}

We integrate our FAU-Net valence score and the pre-trained scoring model with a simple linear combination:
\begin{equation}
    s_{m} + a_{m}s_\mathrm{FAU},
    \label{eqn:combine_pretrained_and_FAU-Net}
\end{equation}
where $m$ represents the specific pre-trained scoring model, $s_{m}$  the score from model $m$, and $a_{m}$ the weight for integrating $m$ and the FAU-Net valence score.

We separately investigated the effect of integrating the FAU-Net valence score with each of the three scoring models trained on large human preference datasets (ImageReward, PickScore, and HPS v2), and their ensemble. Here, we only fit the weights $a_\mathrm{IR}$, $a_\mathrm{Pick}$, $a_\mathrm{HPSv2}$, and $a_\mathrm{ens}$ individually with an LOPO procedure using grid search (from 0.0 to 10.0 with a granularity of 0.1); the FAU-Net weights and the weightings $w_\mathrm{IR}$, $w_\mathrm{Pick}$, $w_\mathrm{HPSv2}$ computed in \Cref{eqn:define_ensemble_baseline_score}
are not modified at this stage. 

\begin{table}[tb]
  \caption{Accuracy of image preference prediction for a total of 3734 image pairs. The second column shows the accuracy of the models given in the first column when they make predictions independently while the third column shows the accuracy of the best few when they are integrated with the FAU-Net valence score.}
  \centering
  \begin{tabular}{ccc}
    \hline 
    \textbf{Model} & \textbf{Ind. Acc.} & \textbf{Acc. w/ AUs} \\
    \hline
    CLIP Score & 57.74 & / \\
    \hline
    Aesthetic Score & 50.32 & /\\
    \hline
    BLIP Score & 50.05 & /\\
    \hline
    FAU-Net Valence Score & 59.35 & /\\
    \hline
    ImageReward Score & 60.47 & 65.21\\
    \hline
    PickScore & 63.28 & 66.28\\
    \hline
    HPS V2 Score & 62.00 & 66.87\\
    \hline
    \textbf{Ensemble Baseline Score} & 65.80 & \textbf{68.64}\\
    \hline
  \end{tabular}
  \label{tab:accuracy_of_independent_and_integrated_predictions}
\end{table}

The results are shown in \Cref{tab:accuracy_of_independent_and_integrated_predictions}. Integration with FAU-Net valence score improves the performance of all four baseline models. Specifically, by further integrating  all three pre-trained human preference scoring models together with the FAU-Net valence score, we achieve the highest accuracy of 68.64\%, outperforming the ensemble of the three models by a margin of 2.84\% ($p<0.01$ with a two-proportion z-test).

\subsubsection{Complementarity between FAU-Net and Pre-trained Scoring Models}
\begin{figure*}[tb]
  \centering
  \begin{subfigure}{0.19\linewidth}
    \includegraphics[width=\linewidth]{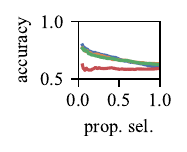}
    \caption{Selected by\\ ImageReward score}
    \label{fig:accuracy_of_data_selected_by_ImageReward_Score}
  \end{subfigure}
  \centering
  \begin{subfigure}{0.19\linewidth}
    \includegraphics[width=\linewidth]{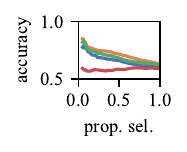}
    \caption{Selected by\\ PickScore}
    \label{fig:accuracy_of_data_selected_by_PickScore}
  \end{subfigure}
  \centering
  \begin{subfigure}{0.19\linewidth}
    \includegraphics[width=\linewidth]{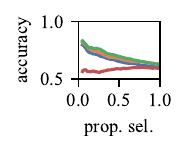}
    \caption{Selected by\\ HPS V2 Score}
    \label{fig:accuracy_of_data_selected_by_HPS_V2_Score}
  \end{subfigure}
  \centering
  \begin{subfigure}{0.41\linewidth}
    \includegraphics[width=\linewidth]{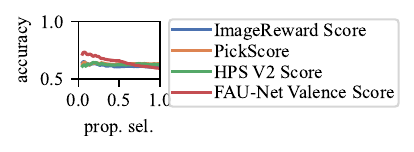}
    \caption{Selected by\\ FAU-Net Valence Score}
    \label{fig:accuracy_of_data_selected_by_FAU-Net_Valence_Score}
  \end{subfigure}
  \caption{\textbf{FAU-Net valence score is complementary to the other pre-trained scoring models.} Each subfigure shows how annotation accuracy changes if only a subset of the data are selected to annotate by the different scoring models (given in the subcaption label), with the x-axis representing the proportion of data selected (image pairs are selected based on the largest absolute difference of  scores for the two images)  and the y-axis representing the annotation accuracy within the selected subset.  Each model is good at estimating the image pairs that it will perform well on as  evidenced by increased accuracy for a smaller selected proportion. Note the blue, green, and orange curves are almost on top of each other.}
  \label{fig:accuracy_of_selected_data}
\end{figure*}

The complementarity between FAU-Net and pre-trained scoring models is demonstrated in \Cref{fig:accuracy_of_selected_data}. The figure shows how annotation accuracy changes if only a subset of the data are selected to annotate by the different scoring models (image pairs are selected based on the largest absolute difference scores for the two images given by the associated scoring model). We can clearly see that the three pre-trained scoring models are highly correlated showing similar performance curves (almost overlapping). In contrast, our FAU-Net valence score selects different image pairs, explaining why the FAU-Net valence scores combine synergistically with the other scores.

To better understand the technical details, let’s first take \Cref{fig:accuracy_of_data_selected_by_FAU-Net_Valence_Score} as an example. In \Cref{fig:accuracy_of_data_selected_by_FAU-Net_Valence_Score}, we first rank the image pairs in the dataset by the absolute value of the difference in FAU-Net scores for the two images, with the assumption that larger score difference implies higher confidence in predicting that the user prefers the image with higher score in the pair and thus higher accuracy. The x-axis of the figure represents ``proportion selected'', which is the proportion of ``most confident'' image pairs—with left side meaning only the image pairs with highest confidence taken into account while right side meaning all image pairs taken into account. The red line in the figure represents the accuracies of FAU-Net for different proportions of ``most confident'' image pairs selected, and indeed as expected, the accuracy is clearly higher on the left side for the most confident image pairs, which confirms that our model is well calibrated. Now note that on the left side (low proportion of data selected) of the figure, the red line is obviously higher than the other lines, representing other pre-trained scoring models. This implies that for the image pairs with largest differences in their FAU-Net scores, the choice made by the FAU-Net is significantly more accurate than the choice made by the other models—in other words, for the image pairs our FAU-Net model is very good at, the other models are not good at them.

At first glance, this might seem like a matter of course and not impressive, but this is actually not true for other models. Let’s then take \Cref{fig:accuracy_of_data_selected_by_ImageReward_Score} as an example. As shown in its subcaption in \Cref{fig:accuracy_of_data_selected_by_ImageReward_Score}, the ``most confident image pairs to compare'' here are selected by choosing the image pairs with largest differences in their ImageReward scores. We can see that the ImageReward score (represented by the blue line) is well calibrated, but is also almost overlapping with the orange line and the green line—they are highly correlated. This implies that for the image pairs the ImageReward model is very good at, the PickScore and the HPS V2 models are also similarly very good at, so they are less complementary to each other. We can see similar results in \Cref{fig:accuracy_of_data_selected_by_HPS_V2_Score,fig:accuracy_of_data_selected_by_PickScore} as well. Overall in all of \Cref{fig:accuracy_of_data_selected_by_ImageReward_Score,fig:accuracy_of_data_selected_by_HPS_V2_Score,fig:accuracy_of_data_selected_by_PickScore,fig:accuracy_of_data_selected_by_FAU-Net_Valence_Score}, the curves for all three pre-trained scoring models are highly correlated while our FAU-Net score is not correlated with them.

This demonstrates the uniqueness of our FAU-Net valence score, being very complementary to the other models. We can think about how it works as follows. For the image pairs with small differences in FAU-Net scores (for instance if a user does not show much facial expression difference), the FAU-Net scores will not affect the final score too much in \Cref{eqn:combine_pretrained_and_FAU-Net}, so the other models dominate the decision. However, for the image pairs with large differences in FAU-Net scores, the FAU-Net scores dominate the final score in \Cref{eqn:combine_pretrained_and_FAU-Net}, so our final choice decision is now more based on the FAU-Net model than the other models, and these are exactly the image pairs where our FAU-Net score has much better accuracy than the other models.

\definecolor{darkgreen}{rgb}{0.0, 0.7, 0.0}
\begin{figure*}[tb]
  \centering
  \begin{subfigure}{0.24\linewidth}
    \includegraphics[width=\linewidth]{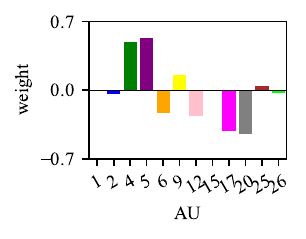}
    \caption{${\color{red} w_{h(1)-o}\approx -0.58}$}
    \label{fig:node_1_weights}
  \end{subfigure}
  \centering
  \begin{subfigure}{0.24\linewidth}
    \includegraphics[width=\linewidth]{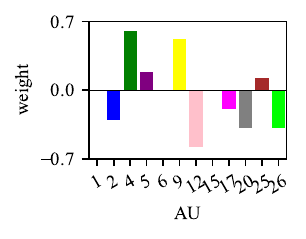}
    \caption{${\color{red} w_{h(2)-o}\approx -0.53}$}
    \label{fig:node_2_weights}
  \end{subfigure}
  \centering
  \begin{subfigure}{0.24\linewidth}
    \includegraphics[width=\linewidth]{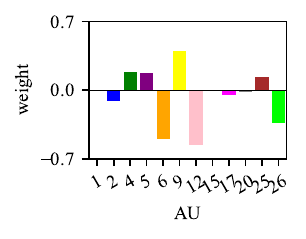}
    \caption{${\color{red} w_{h(3)-o}\approx -0.50}$}
    \label{fig:node_3_weights}
  \end{subfigure}
  \centering
  \begin{subfigure}{0.24\linewidth}
    \includegraphics[width=\linewidth]{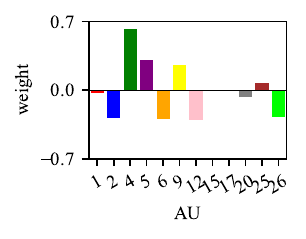}
    \caption{${\color{red} w_{h(4)-o}\approx -0.49}$}
    \label{fig:node_4_weights}
  \end{subfigure}
  \centering
  \begin{subfigure}{0.24\linewidth}
    \includegraphics[width=\linewidth]{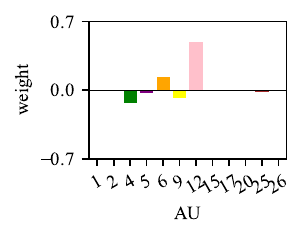}
    \caption{${\color{darkgreen} w_{h(13)-o}\approx 0.30}$}
    \label{fig:node_13_weights}
  \end{subfigure}
  \centering
  \begin{subfigure}{0.24\linewidth}
    \includegraphics[width=\linewidth]{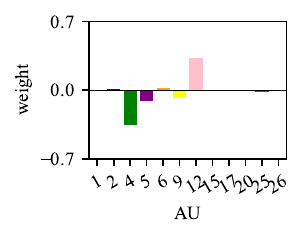}
    \caption{${\color{darkgreen} w_{h(14)-o}\approx 0.31}$}
    \label{fig:node_14_weights}
  \end{subfigure}
  \centering
  \begin{subfigure}{0.24\linewidth}
    \includegraphics[width=\linewidth]{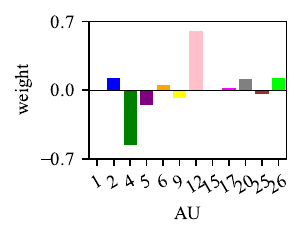}
    \caption{${\color{darkgreen} w_{h(15)-o}\approx 0.42}$}
    \label{fig:node_15_weights}
  \end{subfigure}
  \centering
  \begin{subfigure}{0.24\linewidth}
    \includegraphics[width=\linewidth]{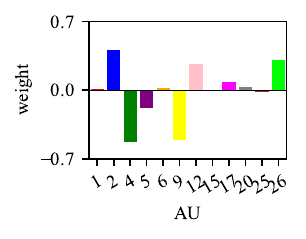}
    \caption{${\color{darkgreen} w_{h(16)-o}\approx 0.56}$}
    \label{fig:node_16_weights}
  \end{subfigure}

  \caption{\textbf{Weights of hidden nodes.} Subfigures show the weights from the input preprocessed activation values of the 12 AUs to a hidden node in the FAU-Net, with each subcaption giving the weight from that hidden node to the output node (red/green for negative/positive weights). Four hidden nodes with lowest/highest negative/positive weights to the output node  are shown here. Full version  with all 16 hidden nodes is shown in the supplemental material.}
  \label{fig:weights_of_hidden_nodes}
\end{figure*}

\begin{figure*}[!htb]
  \centering
    \includegraphics[width=0.23\linewidth]{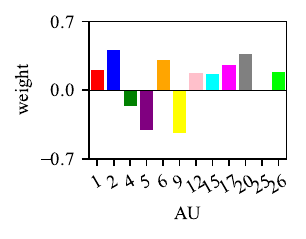}
  \caption{\textbf{Weights of the output node.} The weights from the input preprocessed activation values of the 12 AUs to the output node in the FAU-Net.}
  \label{fig:weights_of_output_node}
\end{figure*}

\subsubsection{FAU-Net Analysis}
We analyze the weights of the trained FAU-Net to better understand the model. \Cref{fig:weights_of_hidden_nodes} shows the weights from the input preprocessed AU activation values to the nodes in the hidden layer (given that the preprocessings of activation values are still all increasing functions after model training) with the subcaptions illustrating the weight from the corresponding hidden node to the output node. Generally, for hidden nodes positively affecting the output score, they receive positive weights from the activations of AU2, AU6, AU12, AU17, AU20, and AU26 and negative weights from the activations of AU4, AU5, AU9, AU17, and AU25, and for hidden nodes negatively affecting the output score, they receive weights of the opposite signs from the AUs. Similar trends are observed for the weights directly from the input preprocessed AU activation values and the output node as shown in \Cref{fig:weights_of_output_node}.
\section{Discussion and Conclusion}

We propose that user facial expression reactions to text-to-image generation can be used to score user preferences for the generated images, which is complementary to existing scoring models which estimate human preferences based only on the input text prompts and generated images. We present the FERGI dataset, consisting of video recordings of facial expression reaction to text-to-image generation, and show that multiple AUs are correlated with the participant's evaluation of and emotional reaction to the generated image.

Specifically, we developed an FAU-Net that receives the detected AU activations in the facial expression reaction as inputs and outputs an FAU-Net valence score as an estimation of user preference. Integrating the FAU-Net valence score with pre-trained scoring models improves their accuracy in image preference prediction, which can be potentially helpful when the scoring models are used to label human preferences for fine-tuning text-to-image generative models.  

Importantly,
previous work has shown that even if the accuracy of the predictions is not very high, it can still be effective for fine-tuning of text-to-image models: \cite{xu2023imagereward} used ImageReward (trained from manually labeled human preferences), with only an accuracy of 65.14\% on their own dataset and 60.47\% on our FERGI dataset, to fine-tune SD v1.4 and obtained a new model whose generated images were preferred by humans 58.79\% of the time over those generated by the original SD v1.4. (see Tab. 4 and Fig. 6 of \cite{xu2023imagereward}).

We also demonstrated the complementarity between FAU-Net and the pre-trained scoring models, explaining why integrating them effectively improves the annotation accuracy. 

Admittedly, our study has some limitations: (1) users may be reluctant to turn on their cameras in real life; (2) the participants of our study are explicitly aware that their reaction videos are being recorded, which might affect the spontaneousness of their facial expressions \cite{zeng2007survey,nichols2008good}.

In conclusion, we have demonstrated the feasibility of automatically scoring user preferences for image generation from facial expression reaction. The application of this method may also be potentially generalizable to other image generation tasks, such as image-to-image translation \cite{pang2021image}, image inpainting \cite{elharrouss2020image}, and super-resolution \cite{wang2020deep}.
\section{Acknowledgments}

We thank Xiaojing Xu and Yuan Tang for helpful prior work, Gary Cottrell, Vijay Veerabadran, and Miguel Monares for helpful discussions, and reviewers for helpful comments. We are grateful for support from NSF IIS 1817226 and CRCNS 2208362, seed funding from UC San Diego Social Sciences and the Sanford Institute for Empathy and Compassion, and gift support from an anonymous donor, as well as hardware funding from NVIDIA, Adobe, and Sony.

\FloatBarrier
{
    \small
    \bibliographystyle{ieeenat_fullname}
    \bibliography{main}
}

\setcounter{section}{0}
\setcounter{table}{0}
\setcounter{figure}{0}
\setcounter{footnote}{0}

\renewcommand{\thesection}{\Alph{section}}
\renewcommand{\thetable}{S\arabic{table}}
\renewcommand{\thefigure}{S\arabic{figure}}

\clearpage
\setcounter{page}{1}
\maketitlesupplementary

\section{Supplementary Visuals}

\begin{figure*}[tb]
  \begin{subfigure}{0.24\linewidth}
    \centering
    \includegraphics[width=0.95\linewidth]{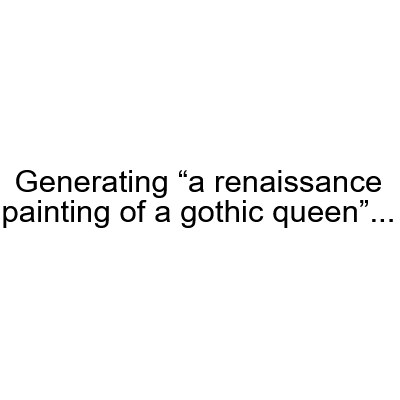}
    \caption{}
    \label{fig:example_prompt_AU4}
  \end{subfigure}
  \begin{subfigure}{0.24\linewidth}
    \centering
    \includegraphics[width=0.8\linewidth]{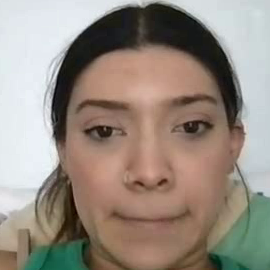}
    \caption{}
    \label{fig:example_baseline_AU4}
  \end{subfigure}
  \begin{subfigure}{0.24\linewidth}
    \centering
    \includegraphics[width=0.8\linewidth]{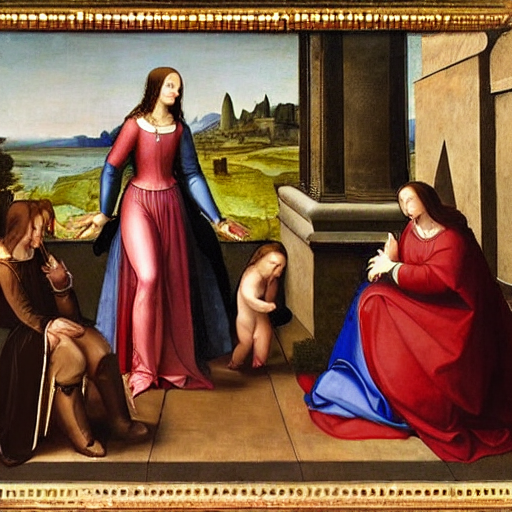}
    \caption{}
    \label{fig:example_image_AU4}
  \end{subfigure}
  \begin{subfigure}{0.24\linewidth}
    \centering
    \includegraphics[width=0.8\linewidth]{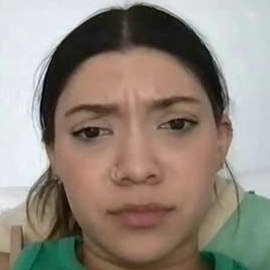}
    \caption{}
    \label{fig:example_reaction_AU4}
  \end{subfigure}\\
  \caption{\textbf{An example of AU4 activation in response to a low-quality image generation.} 
  (a) The input text prompt.
  (b) The facial expression of the participant before seeing the generated image.
  (c) The generated image.
  (d) The facial expression of the participant after seeing the generated image (the frame with the highest estimated
  AU4 intensity).
  In the annotation survey, the participant gave an overall rating of 2, an image-text alignment rating of 2, and a fidelity rating of 1; she identified the image with the issues ``output contains unwanted content that was not mentioned in the text prompt'' and ``existence of body problem''; she reported that she felt ``disappointed'' when seeing the image; she ranked the image as the worst among the five images generated from the same prompt.}
  \label{fig:example_AU4}
\end{figure*}

\begin{figure*}[tb]
  \begin{subfigure}{0.24\linewidth}
    \centering
    \includegraphics[width=0.95\linewidth]{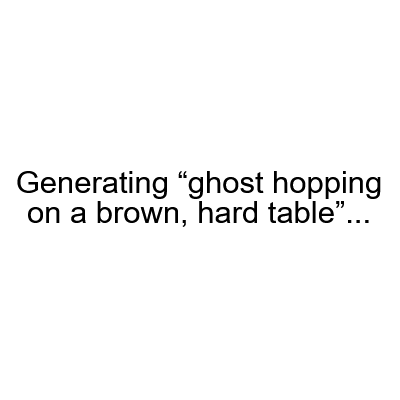}
    \caption{}
    \label{fig:example_prompt_AU12}
  \end{subfigure}
  \begin{subfigure}{0.24\linewidth}
    \centering
    \includegraphics[width=0.8\linewidth]{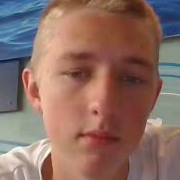}
    \caption{}
    \label{fig:example_baseline_AU12}
  \end{subfigure}
  \begin{subfigure}{0.24\linewidth}
    \centering
    \includegraphics[width=0.8\linewidth]{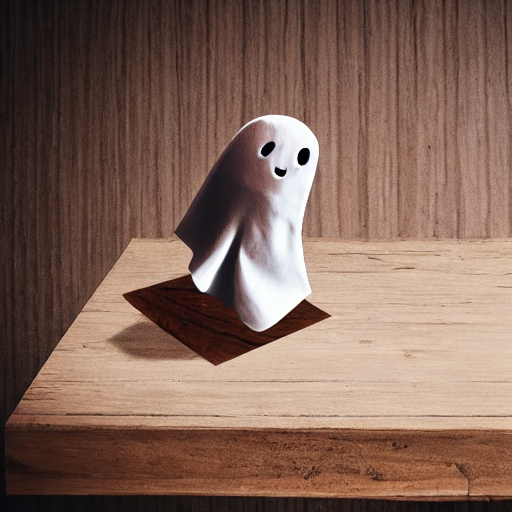}
    \caption{}
    \label{fig:example_image_AU12}
  \end{subfigure}
  \begin{subfigure}{0.24\linewidth}
    \centering
    \includegraphics[width=0.8\linewidth]{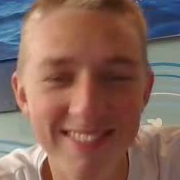}
    \caption{}
    \label{fig:example_reaction_AU12}
  \end{subfigure}
  \caption{\textbf{An example of AU12 activation in response to a high-quality image generation.} 
  (a) The input text prompt.
  (b) The facial expression of the participant before seeing the generated image.
  (c) The generated image.
  (d) The facial expression of the participant after seeing the generated image (the frame with the highest estimated 
  AU12 intensity).
  In the annotation survey, the participant gave ratings of 7 for each of
image-text alignment, fidelity, and overall rating; he did not identify any issues with the image; he reported feeling ``satisfied'', ``surprised'', and ``amused'' when seeing the image; he ranked the image as the  best among the five images generated from the same prompt.}
  \label{fig:example_AU12}
\end{figure*}

To better illustrate the idea of our paper, we present \Cref{fig:example_AU4} showing an example of a strong AU4 activation in response to a low-quality image generation, and \Cref{fig:example_AU12} an example of a strong AU12 activation in response to a high-quality image generation.

\section{Data Collection}

\begin{figure*}[tb]
  \centering
  \includegraphics[width=.95\linewidth]{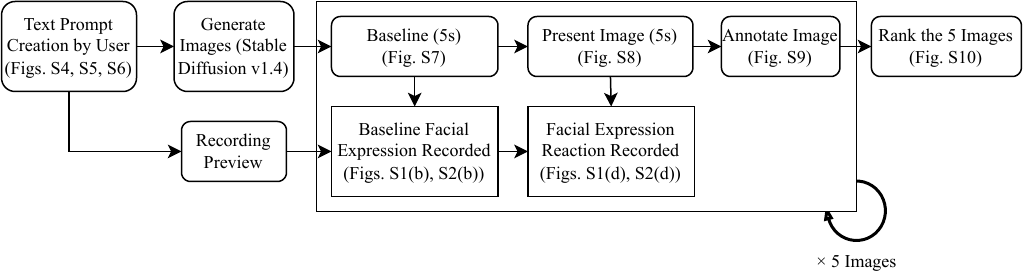}
  \caption{\textbf{Procedure of one session in data collection. A copy of \Cref{fig:data_collection_procedure} with additional references to the screenshots of each stage.} Firstly, the participant creates an input text prompt. Secondly, the participant is directed to a webcam preview to confirm that the webcam captures their face appropriately; at the same time, 5 images are generated from the text prompt using Stable Diffusion v1.4. Then, for each image, the participant goes through a 5-second baseline phase, a 5-second image presentation phase, and an image annotation phase with no time restriction. (During the baseline phase, the participant's baseline facial expression is recorded, and during the image presentation phase, the participant's facial expression reaction to the generated image is recorded.) Finally, after all 5 images have been presented and annotated, the participant ranks the 5 generated images from best to worst.}
  \label{fig:data_collection_procedure_supplemental_material}
\end{figure*}

This section provides more details of the collection of the FERGI dataset not provided in \Cref{sec:FERGI}. \Cref{fig:data_collection_procedure_supplemental_material} displays a copy of \Cref{fig:data_collection_procedure}, showing the procedure of one session in data collection with additional references to the screenshots of each stage. \Cref{fig:basic_inputs_structured_input,fig:prompt_generation_structured_input,fig:prompt_creation_free_form_input,fig:baseline,fig:present,fig:annotate,fig:ranking} display screenshots of an example of one session in data collection.

\subsection{Participants}
39 participants\footnote{The user IDs are up to 040 as shown in \Cref{tab:dataset_info} because the same participant used both 027 and 028 due to some technical issues. Though, no data was collected under the user ID 027.} were recruited from the SONA system of University of California San Diego (UCSD).  and received academic credits as compensation. They underwent the informed consent process and received instruction for the study from the researcher via a videoconference and completed the study asynchronously afterwards on their personal computers. The videos of 3 participants failed to record and displayed only a logo of OBS Studio (the reason has not yet been fully understood by the researchers); 1 participant dropped out of study under the instruction of the researcher because the only available personal device was a tablet, which is not compatible with some procedures of the study; 1 participant did not complete any study sessions asynchronously at all after receiving the instruction in the videoconference. Among the other 34 participants with valid data and videos recorded, only 1 participant chose not to permit video recordings being shared with other researchers for research purposes; therefore, we decided to also exclude the data of this participant in our analysis for enhancing reproducibility of our study within the research community. We ended up with a dataset of 33 participants that will be made available to researchers for research purposes. 

\subsection{Prompt Creation}
There are two forms of sessions, structured input (SI) sessions and free-form input (FFI) sessions.
Most participants completed the same number of structured input sessions and free-form input sessions.
The number of sessions completed by each participant are shown in \Cref{tab:dataset_info}.
By default, each participant is expected to complete 9 structured input sessions and 9 free-form input sessions (for the first 13 out of the 33 participants chronologically) or to complete 10 structured input sessions and 10 free-form input sessions (for the last 20 out of the 33 participants chronologically). However, the time consumption highly varies for each participant because of numerous reasons (e.g. internet connection), and various technical issues also occur occasionally, so the total number and categories of completed sessions also vary.

\begin{table*}[tb]
  \caption{Information of the FERGI dataset. Numbers of structured input (SI) sessions, numbers of free-form input (FFI) sessions, and numbers of valid images for each participant are shown.}
  \centering
  \begin{tabular}{|c|c|c|c|}
    \hline
    User ID & Number of SI Sessions & Number of FFI Sessions & Number of Valid Images\\
    \hline
    001 & 10 & 10 & 100\\
    \hline
    002 & 5 & 5 & 50\\
    \hline
    004 & 9 & 9 & 87\\
    \hline
    005 & 9 & 9 & 85\\
    \hline
    006 & 9 & 9 & 90\\
    \hline
    007 & 9 & 9 & 88\\
    \hline
    008 & 9 & 9 & 85\\
    \hline
    009 & 9 & 9 & 88\\
    \hline
    010 & 9 & 8 & 84\\
    \hline
    011 & 9 & 9 & 87\\
    \hline
    012 & 8 & 8 & 80\\
    \hline
    015 & 9 & 9 & 89\\
    \hline
    016 & 10 & 10 & 100\\
    \hline
    017 & 3 & 0 & 15\\
    \hline
    018 & 11 & 4 & 75\\
    \hline
    020 & 10 & 10 & 92\\
    \hline
    021 & 9 & 9 & 87\\
    \hline
    022 & 9 & 6 & 75\\
    \hline
    023 & 10 & 10 & 99\\
    \hline
    024 & 10 & 2 & 58\\
    \hline
    025 & 3 & 10 & 63\\
    \hline
    026 & 10 & 10 & 100\\
    \hline
    028 & 0 & 20 & 99\\
    \hline
    030 & 10 & 10 & 99\\
    \hline
    031 & 10 & 10 & 100\\
    \hline
    032 & 10 & 10 & 99\\
    \hline
    033 & 10 & 10 & 99\\
    \hline
    034 & 10 & 10 & 98\\
    \hline
    035 & 10 & 10 & 99\\
    \hline
    037 & 10 & 10 & 100\\
    \hline
    038 & 10 & 10 & 96\\
    \hline
    039 & 10 & 10 & 96\\
    \hline
    040 & 9 & 4 & 65\\
    \hline
    Total & 288 & 288 & 2827\\
    \hline
  \end{tabular}
  \label{tab:dataset_info}
\end{table*}

\subsubsection{Structured Input}
\begin{figure*}[tb]
  \centering
  \begin{subfigure}{0.86\linewidth}
    \includegraphics[width=\linewidth]{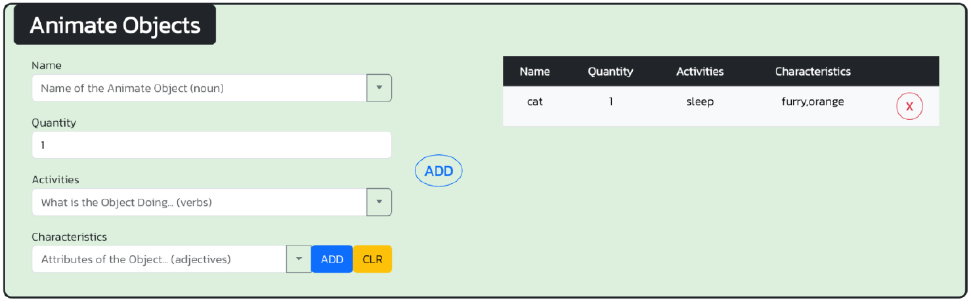}
    \caption{Animate objects}
    \label{fig:animate_obj}
  \end{subfigure}
  \begin{subfigure}{0.86\linewidth}
    \includegraphics[width=\linewidth]{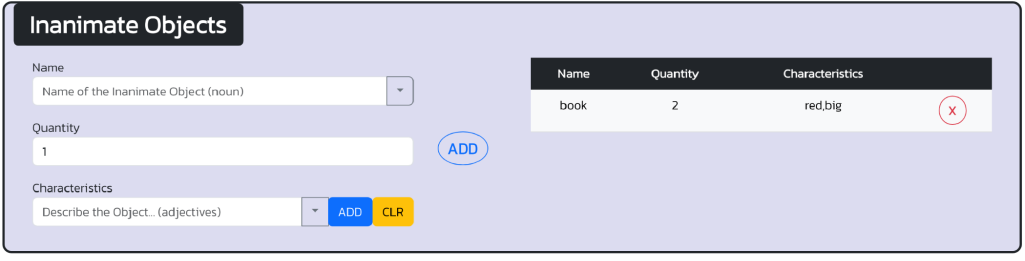}
    \caption{Inanimate objects}
    \label{fig:inanimate_obj}
  \end{subfigure}
  \begin{subfigure}{0.86\linewidth}
    \includegraphics[width=\linewidth]{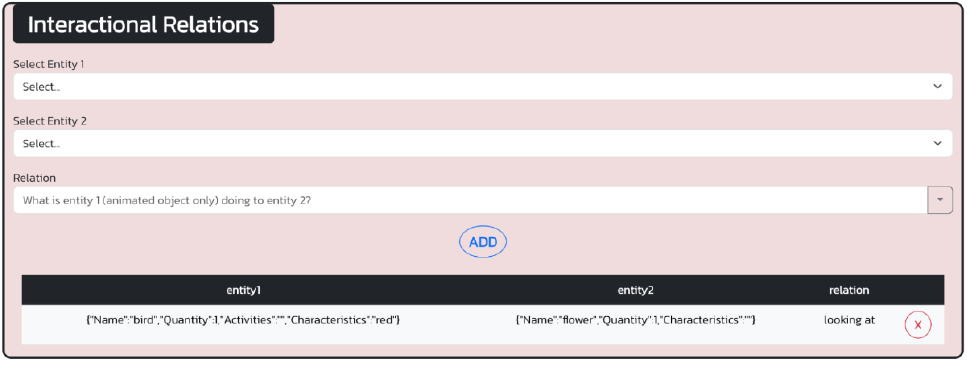}
    \caption{Interactional relations}
    \label{fig:interactional_rel}
  \end{subfigure}
  \begin{subfigure}{0.86\linewidth}
    \includegraphics[width=\linewidth]{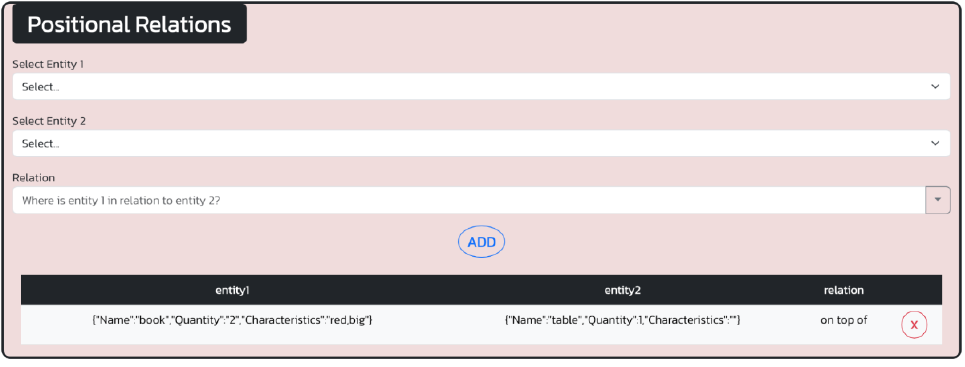}
    \caption{Positional relations}
    \label{fig:positional_rel}
  \end{subfigure}
  \caption{An example of the basic inputs for the prompt creation of structured input sessions.}
  \label{fig:basic_inputs_structured_input}
\end{figure*}

\begin{figure*}[tb]
  \centering
  \includegraphics[width=\linewidth]{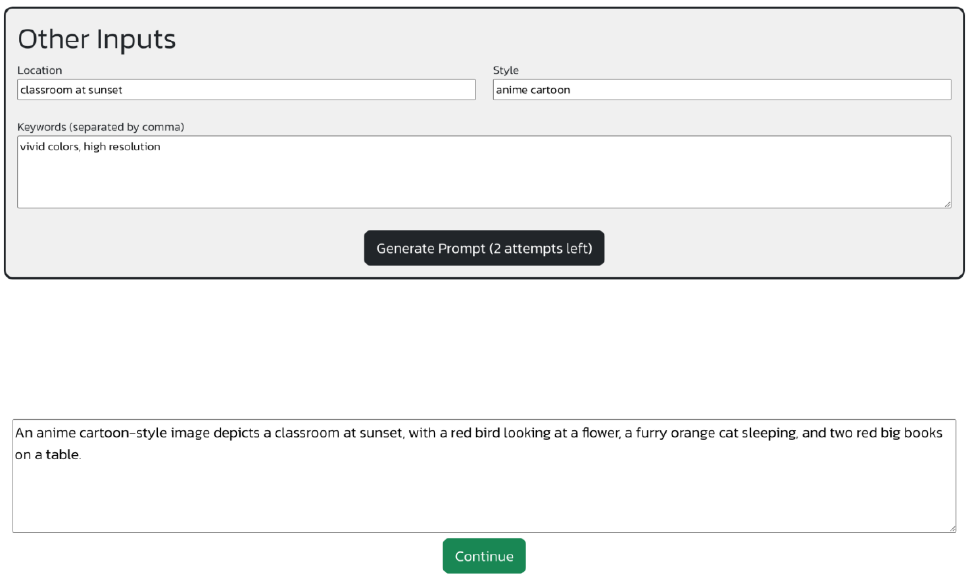}
  \caption{An example of other inputs and prompt generation for the prompt creation of structured input sessions.}
  \label{fig:prompt_generation_structured_input}
\end{figure*}

The purpose of having structured input sessions is to encourage creation of more diverse text prompts featuring different types of objects, interactions, settings, and styles. Structured input contains 5 sections:
\begin{itemize}
\item \textbf{Animate Objects} (\Cref{fig:animate_obj}): Any object that can do things, including humans, animals, robots, fantasy creatures, etc. Participants are required to enter the name and quantity of the object, with activity and a list of characteristics as optional inputs.
\item \textbf{Inanimate Objects} (\Cref{fig:inanimate_obj}): Any object that cannot do things, including plants, vehicles, furniture, etc. Participants are required to enter the name and quantity of the object, with a list of characteristics as optional inputs.
\item \textbf{Interactional Relations} (\Cref{fig:interactional_rel}): Relationship between an animated object and any other object (something an animate object does to another object). An animate object can only be in a relation if its ``activity'' field is empty. All objects can only be involved in at most one relation (either interactional or positional).
\item \textbf{Positional Relations} (\Cref{fig:positional_rel}): Relationship between any two objects. An animate object can only be in a relation if its ``activity'' field is empty. All objects can only be involved in at most one relation (either interactional or positional).
\item \textbf{Other Inputs} (\Cref{fig:prompt_generation_structured_input}): The location of image as background or general environment, the style of the image, and a list of keywords to append at the end of the prompt.
\end{itemize}

After participants finish with the input sections and click ``generate prompt'', a prompt will be generated based on all of the inputs with LLM (OpenAI gpt3.5-turbo model). Participants can freely adjust the generated prompt as the final prompt. Screenshots of an example of the inputs and prompt generation for a structured input session are shown in \Cref{fig:basic_inputs_structured_input,fig:prompt_generation_structured_input}.

\subsubsection{Free-Form Input}

\begin{figure*}[tb]
  \centering
  \includegraphics[width=\linewidth]{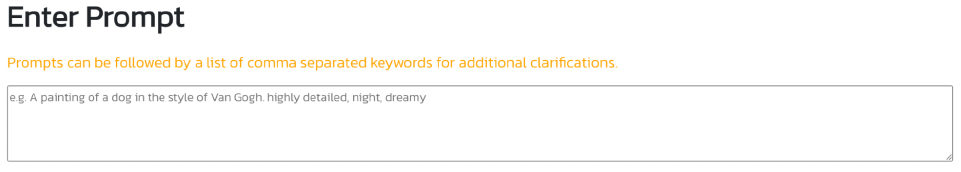}
  \caption{An example of the prompt creation page for the free-form input sessions.}
  \label{fig:prompt_creation_free_form_input}
\end{figure*}

The purpose of free-form input is to provide freedom and flexibility for participants to experiment with any prompts to complement the less flexible input style of structured input. The participants are provided with an example prompt and a text box to enter the entire prompt as the final prompt. A screenshot of an example of the prompt creation page for a free-form input session is shown in \Cref{fig:prompt_creation_free_form_input}.

\subsection{Image Presentation}
\begin{figure*}[tb]
  \centering
  \includegraphics[width=\linewidth]{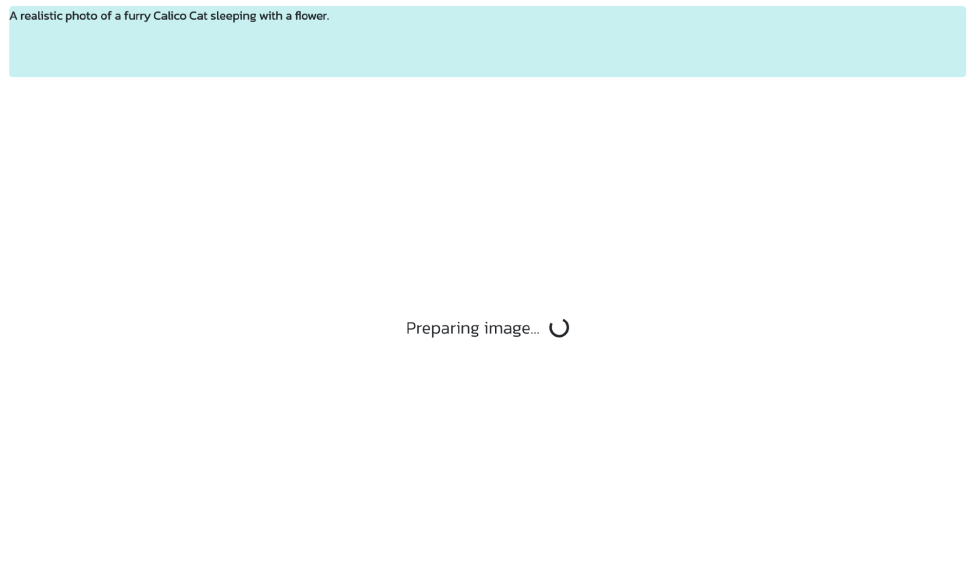}
  \caption{An example of a baseline page.}
  \label{fig:baseline}
\end{figure*}

\begin{figure*}[tb]
  \centering
  \includegraphics[width=\linewidth]{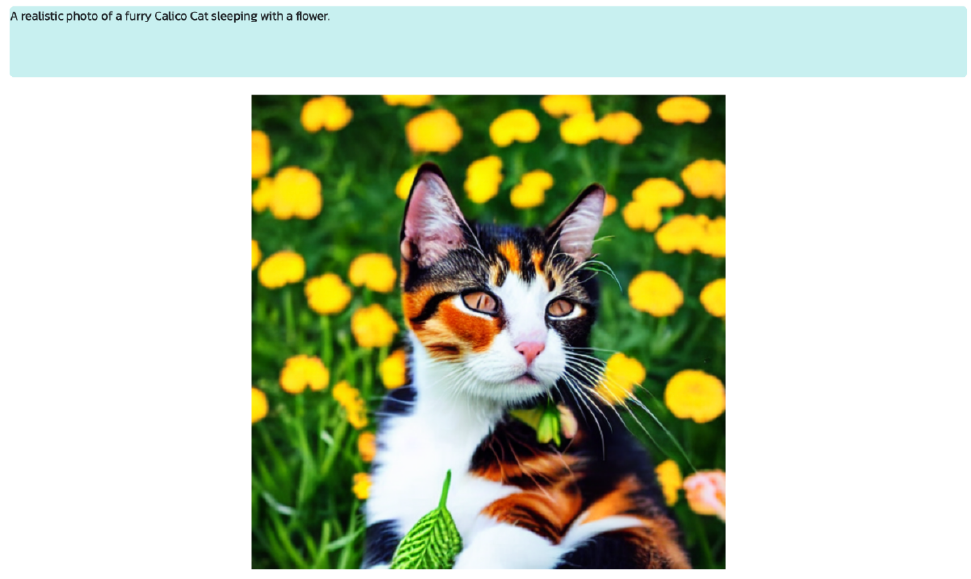}
  \caption{An example of an image presentation page.}
  \label{fig:present}
\end{figure*}

Screenshots of an example of the pages for image presentation and its baseline are shown in \Cref{fig:baseline,fig:present}.

\subsubsection{Image Annotation}
\begin{figure*}[tb]
  \centering
  \includegraphics[width=\linewidth]{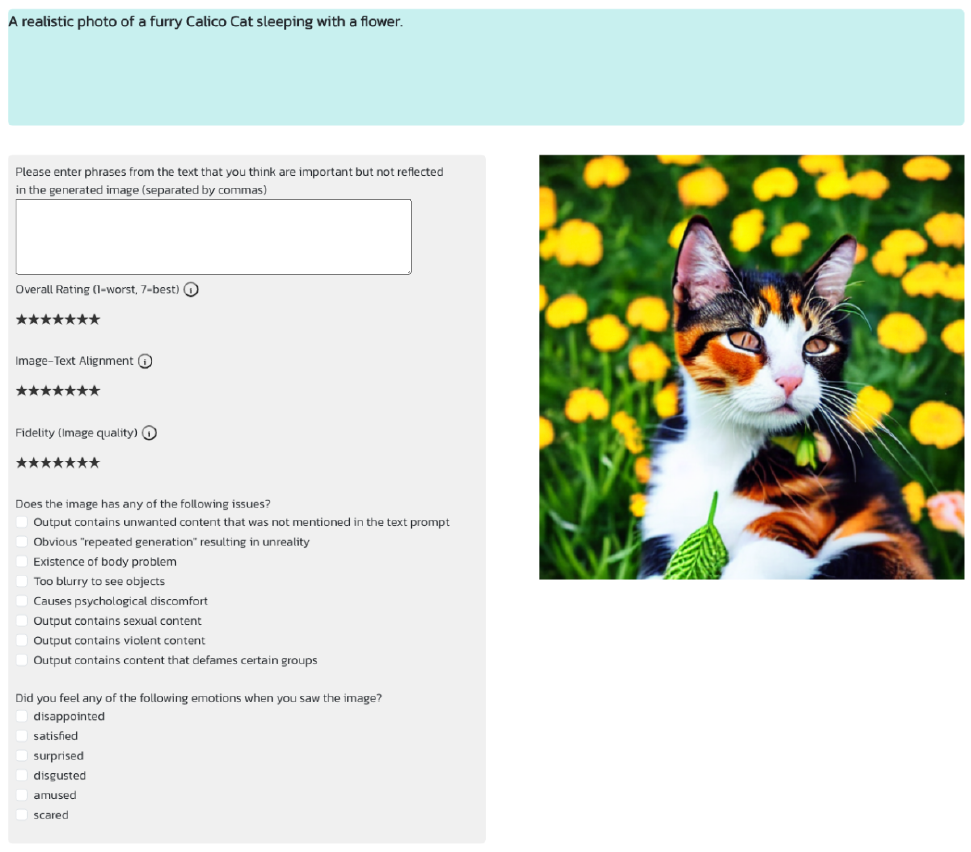}
  \caption{An example of an image annotation survey page.}
  \label{fig:annotate}
\end{figure*}

The annotation survey was adapted from the survey designed for ImageReward \cite{xu2023imagereward} and has the following components:
\begin{itemize}
\item Participants can optionally provide a typed, comma-separated response indicating any phrase or parts of the prompt that are not reflected by the generated image.
\item Participants provide star ratings from 1 to 7 for image-text alignment, fidelity, and overall rating for the generated image.
\item Participants can optionally select from a list of common issues to report the issues the generated image has. 
\item Participants can optionally select from a list of emotions to report their feelings when seeing the image. 
\end{itemize}
A screenshot of an example of the image annotation page (including the details of the questions) is shown in \Cref{fig:annotate}.

\subsubsection{Image Ranking}
\begin{figure*}[tb]
  \centering
  \includegraphics[width=\linewidth]{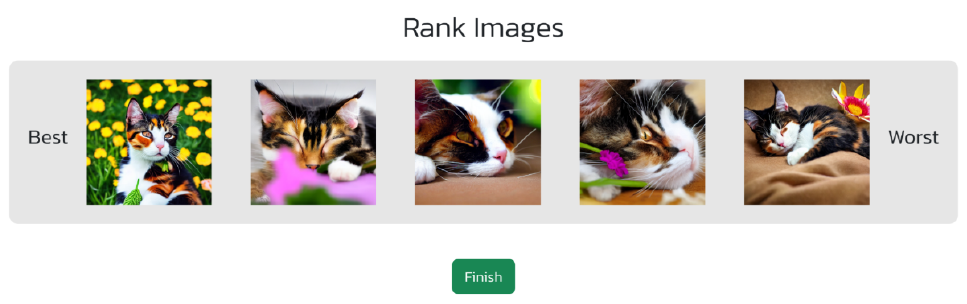}
  \caption{An example of an image ranking page.}
  \label{fig:ranking}
\end{figure*}

After the participants finish annotating all 5 generated images, they will be asked to rank them from best to worst. All 5 images are displayed side by side, with the left side labeled ``Best'' and right side labeled ``Worst'', and the participants provide ranking by dragging the images to reorder them. A screenshot of an example of the image ranking page is shown in \Cref{fig:ranking}.

\section{AU Model Training}
This section provides more details of the training of the AU model not provided in \Cref{sec:ModelTraining}.

\subsection{Datasets}

\begin{figure*}[tb]
  \centering
  \begin{subfigure}{0.4\linewidth}
    \centering
    \includegraphics[width=\linewidth]{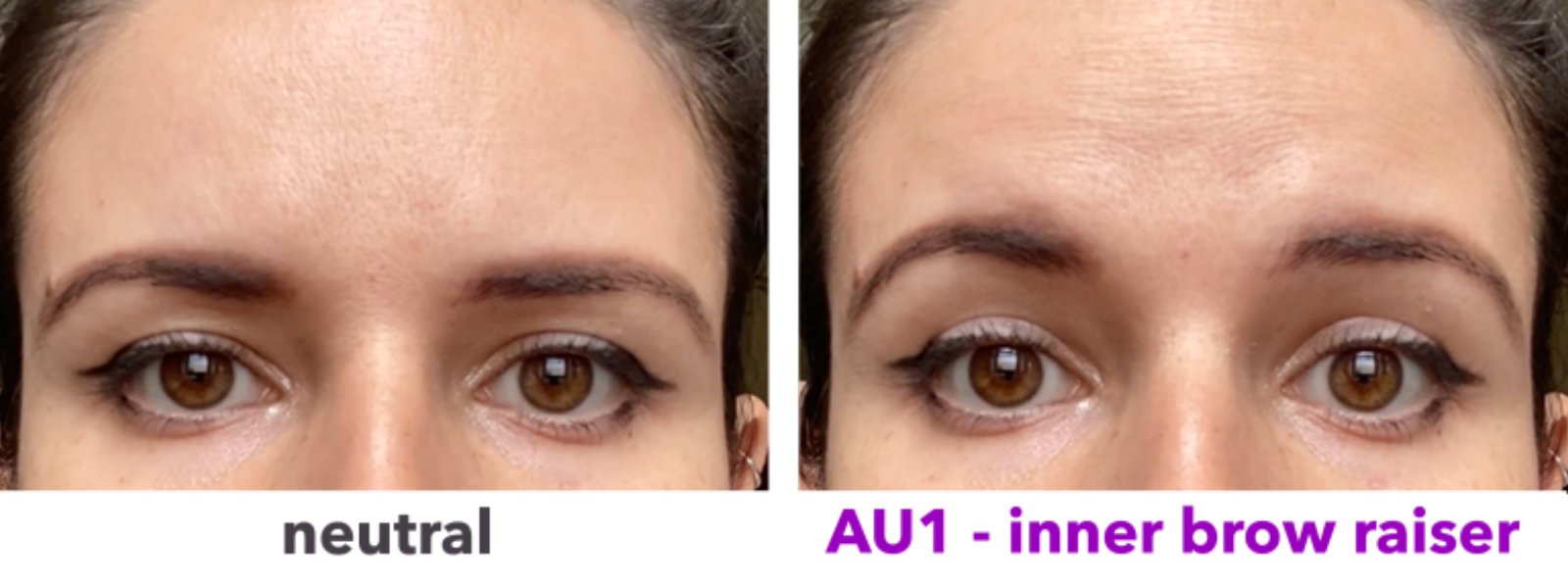}
  \end{subfigure}
  \hspace{0.08\linewidth}
  \begin{subfigure}{0.4\linewidth}
    \centering
    \includegraphics[width=\linewidth]{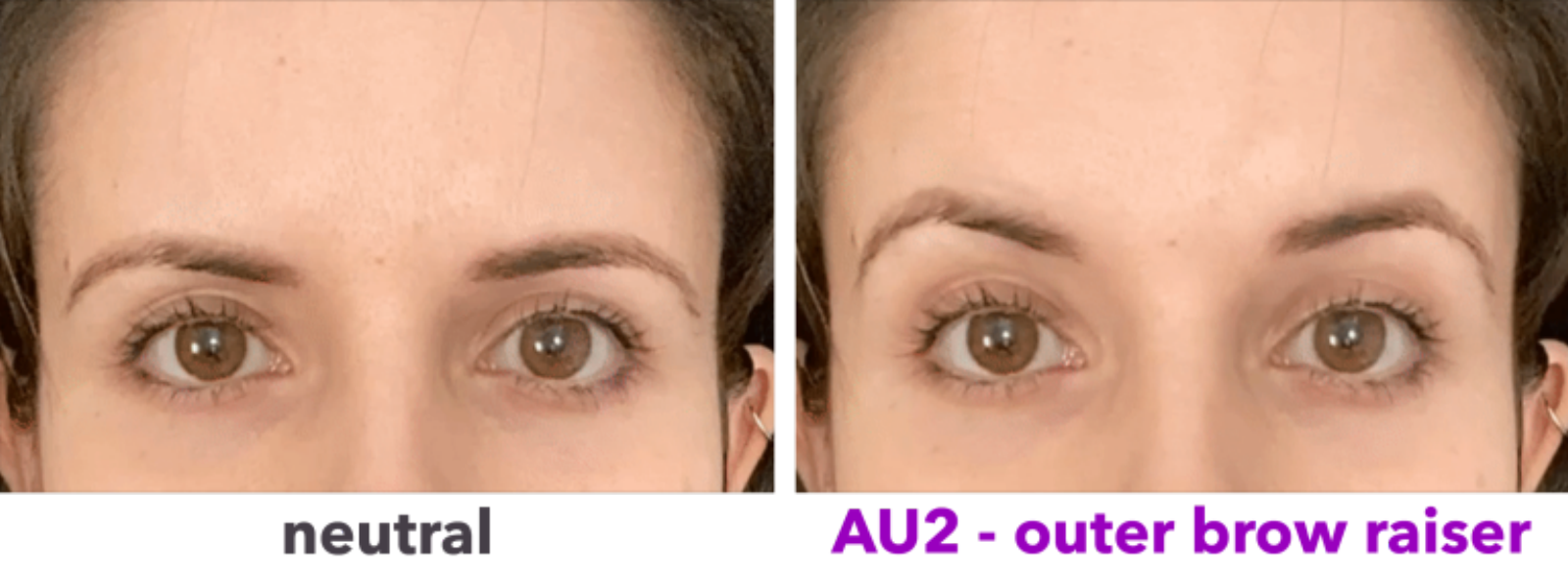}
  \end{subfigure}\\
  \begin{subfigure}{0.4\linewidth}
    \centering
    \includegraphics[width=\linewidth]{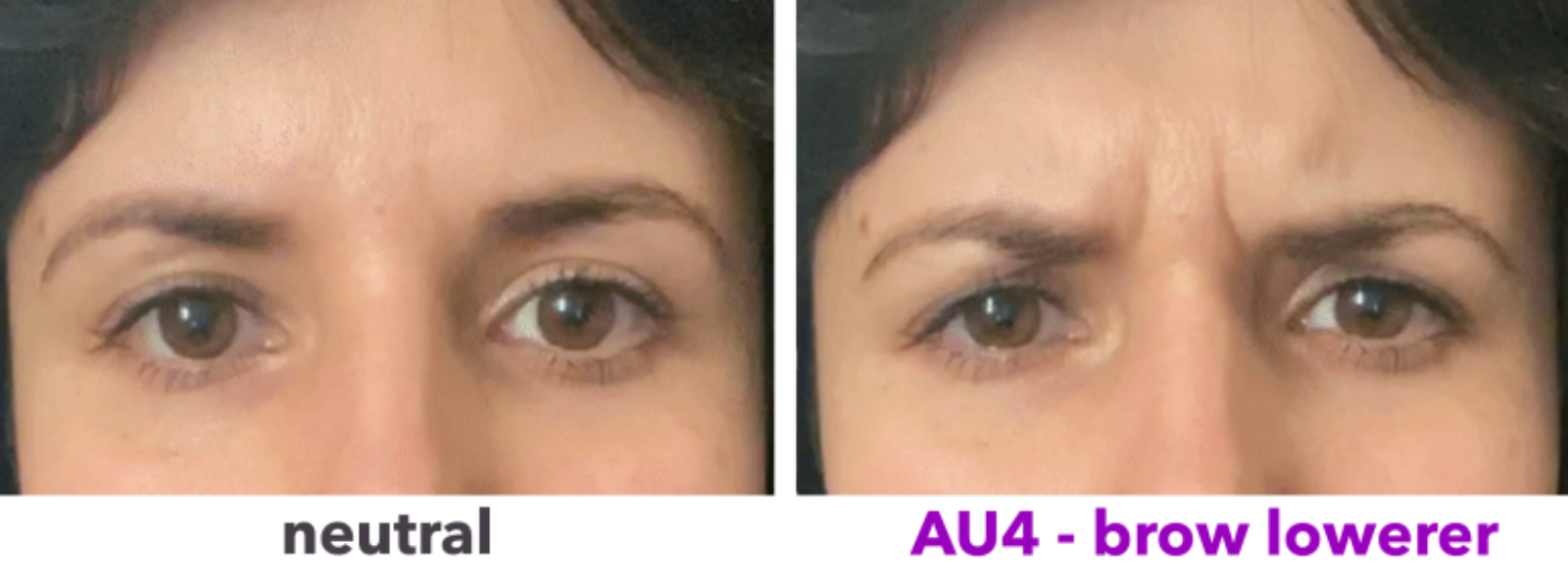}
  \end{subfigure}
  \hspace{0.08\linewidth}
  \begin{subfigure}{0.4\linewidth}
    \centering
    \includegraphics[width=\linewidth]{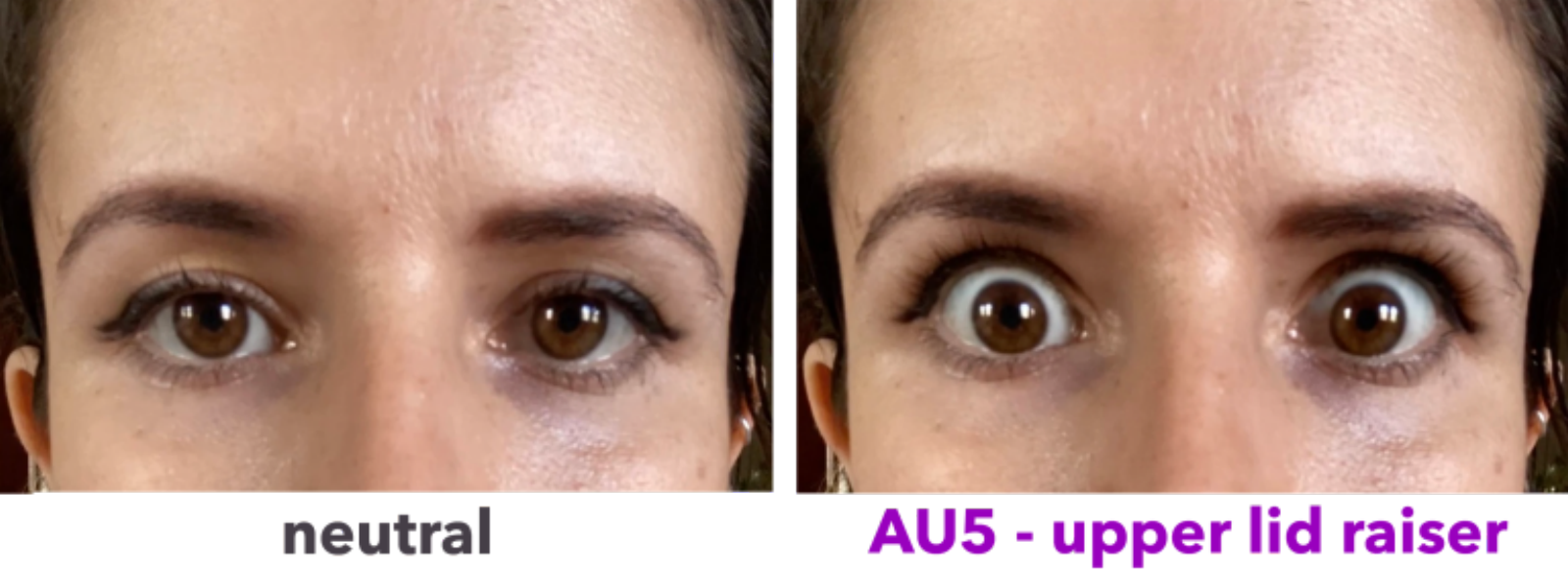}
  \end{subfigure}\\
  \begin{subfigure}{0.4\linewidth}
    \centering
    \includegraphics[width=\linewidth]{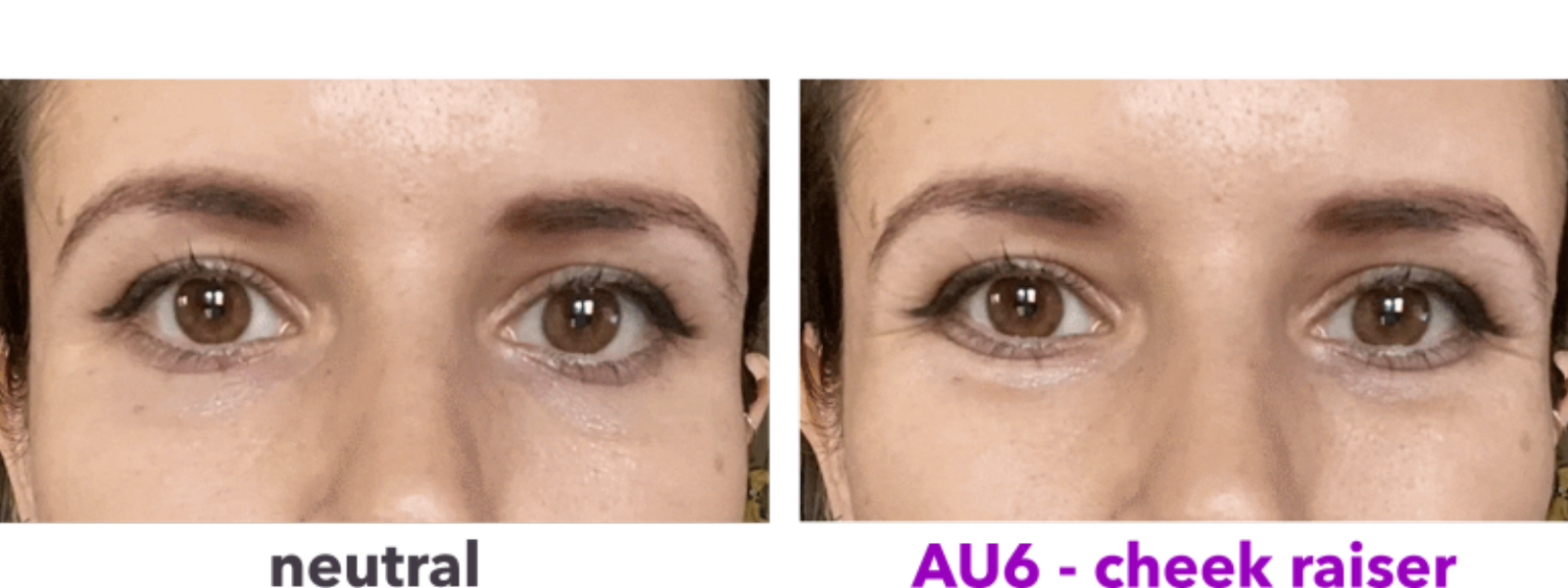}
  \end{subfigure}
  \hspace{0.08\linewidth}
  \begin{subfigure}{0.4\linewidth}
    \centering
    \includegraphics[width=\linewidth]{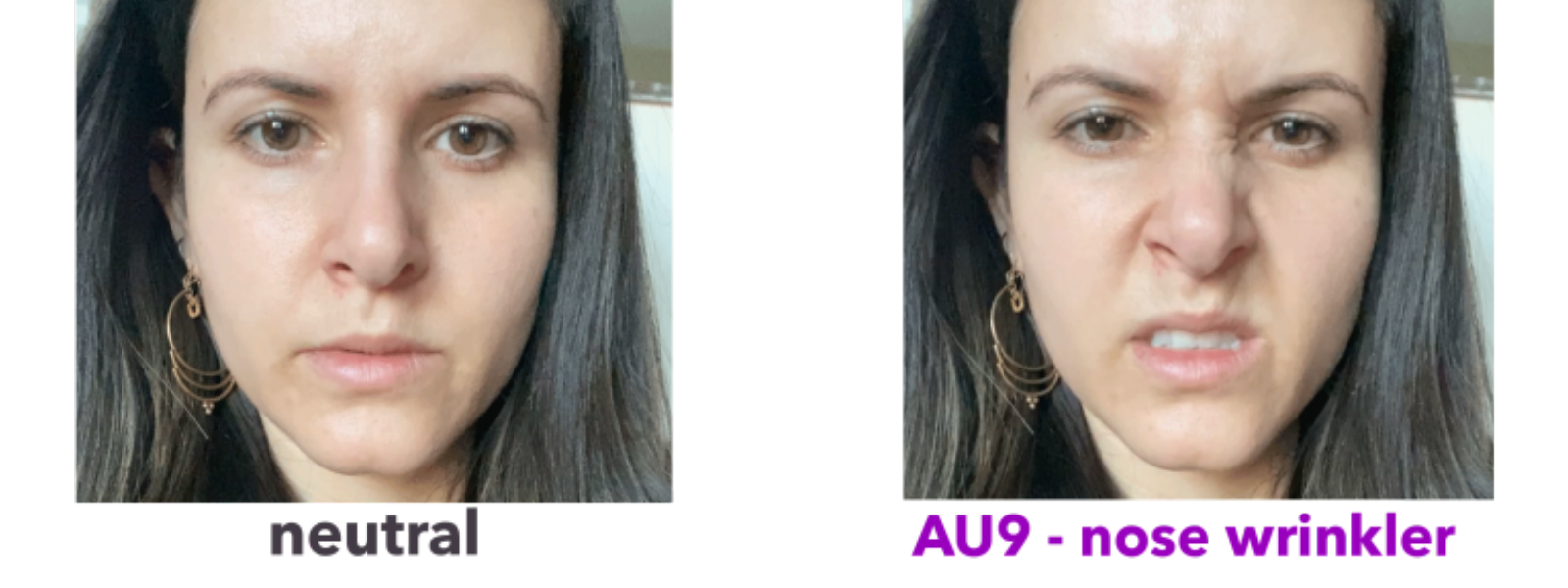}
  \end{subfigure}\\
  \begin{subfigure}{0.4\linewidth}
    \centering
    \includegraphics[width=\linewidth]{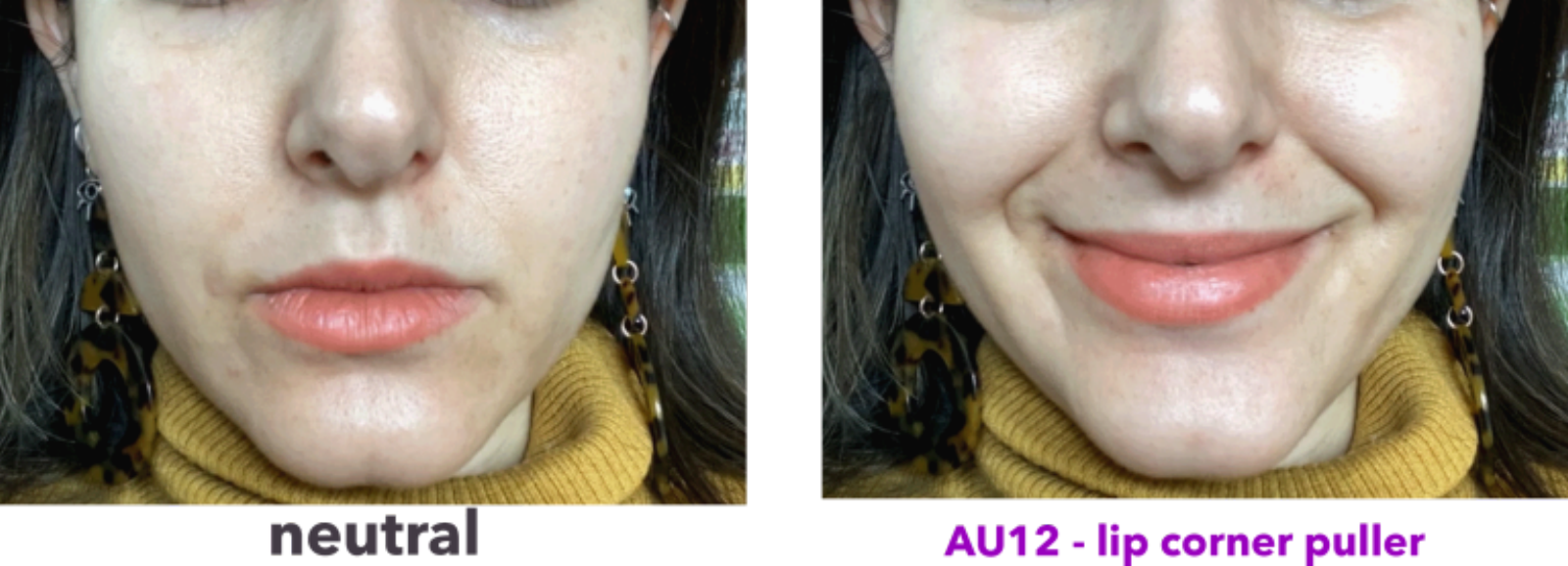}
  \end{subfigure}
  \hspace{0.08\linewidth}
  \begin{subfigure}{0.4\linewidth}
    \centering
    \includegraphics[width=\linewidth]{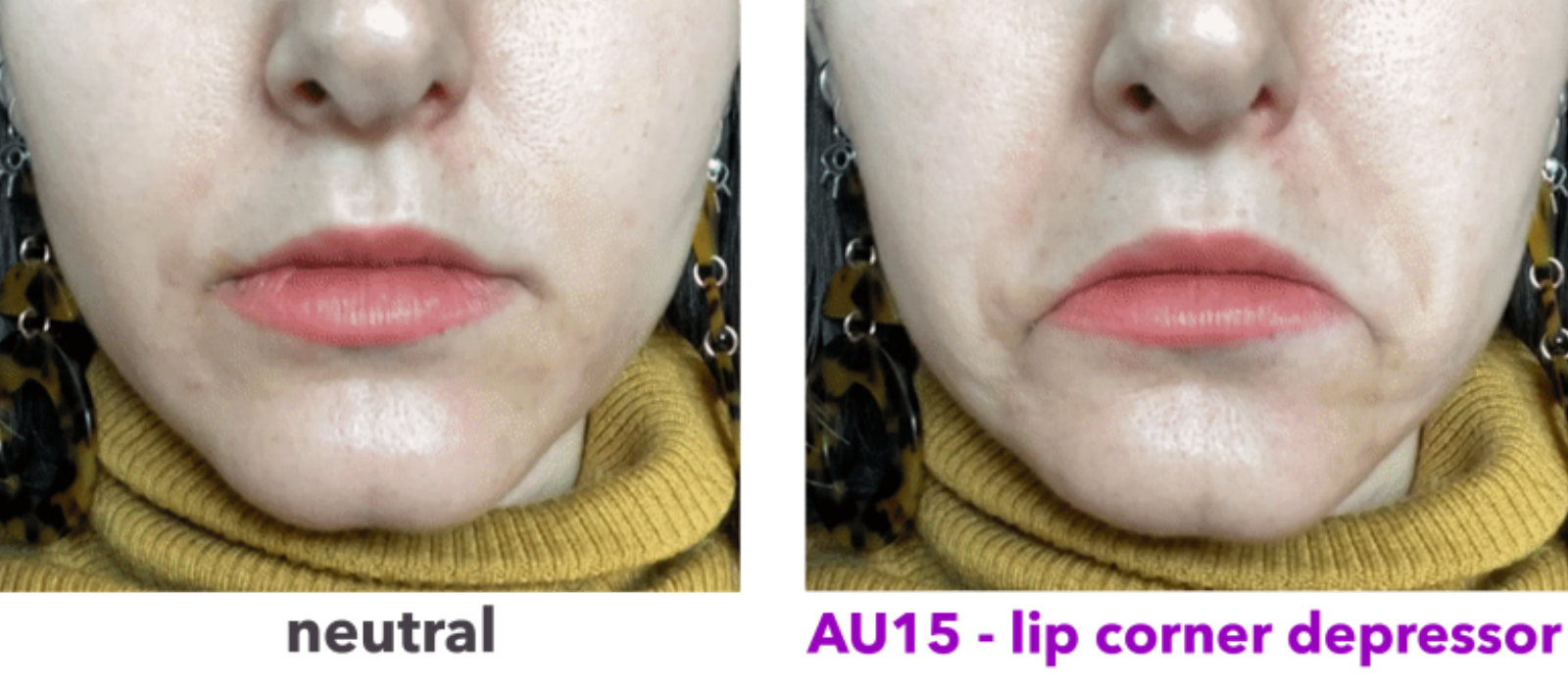}
  \end{subfigure}\\
  \begin{subfigure}{0.4\linewidth}
    \centering
    \includegraphics[width=\linewidth]{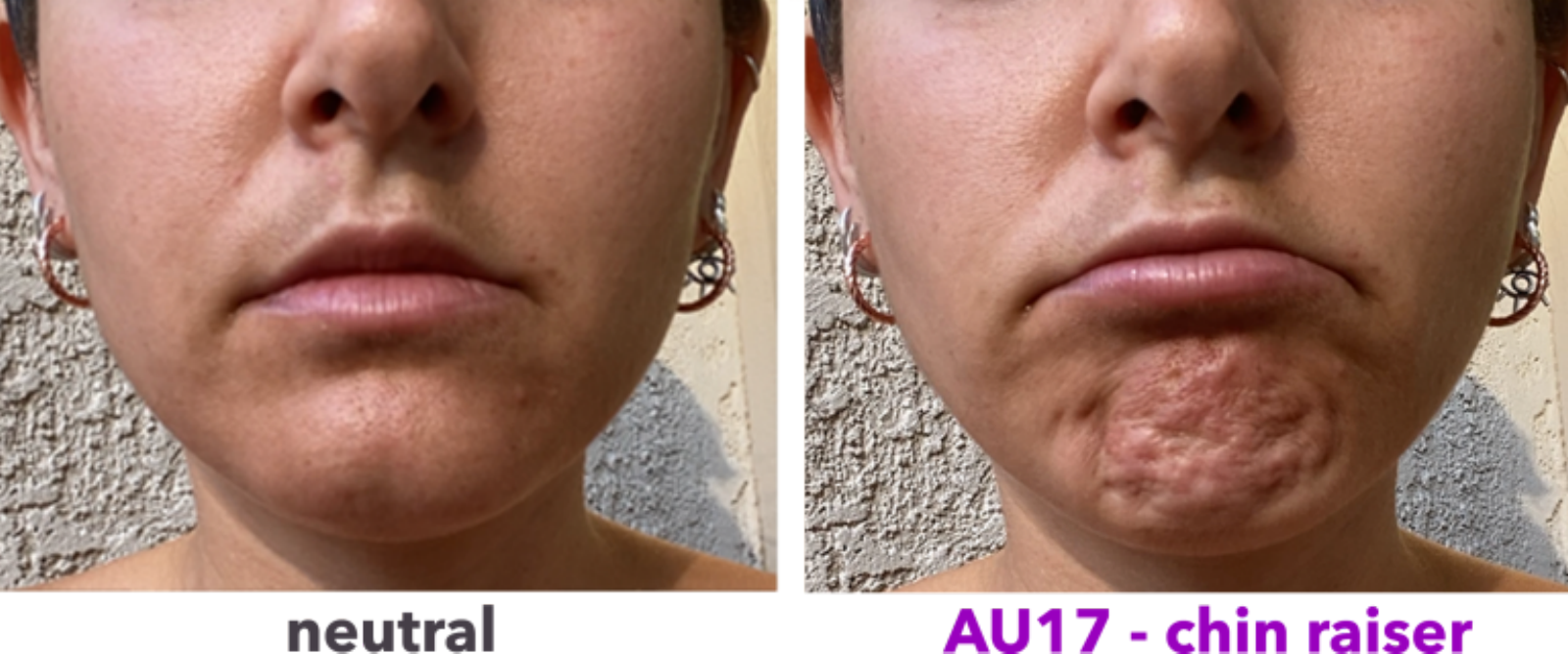}
  \end{subfigure}
  \hspace{0.08\linewidth}
  \begin{subfigure}{0.4\linewidth}
    \centering
    \includegraphics[width=\linewidth]{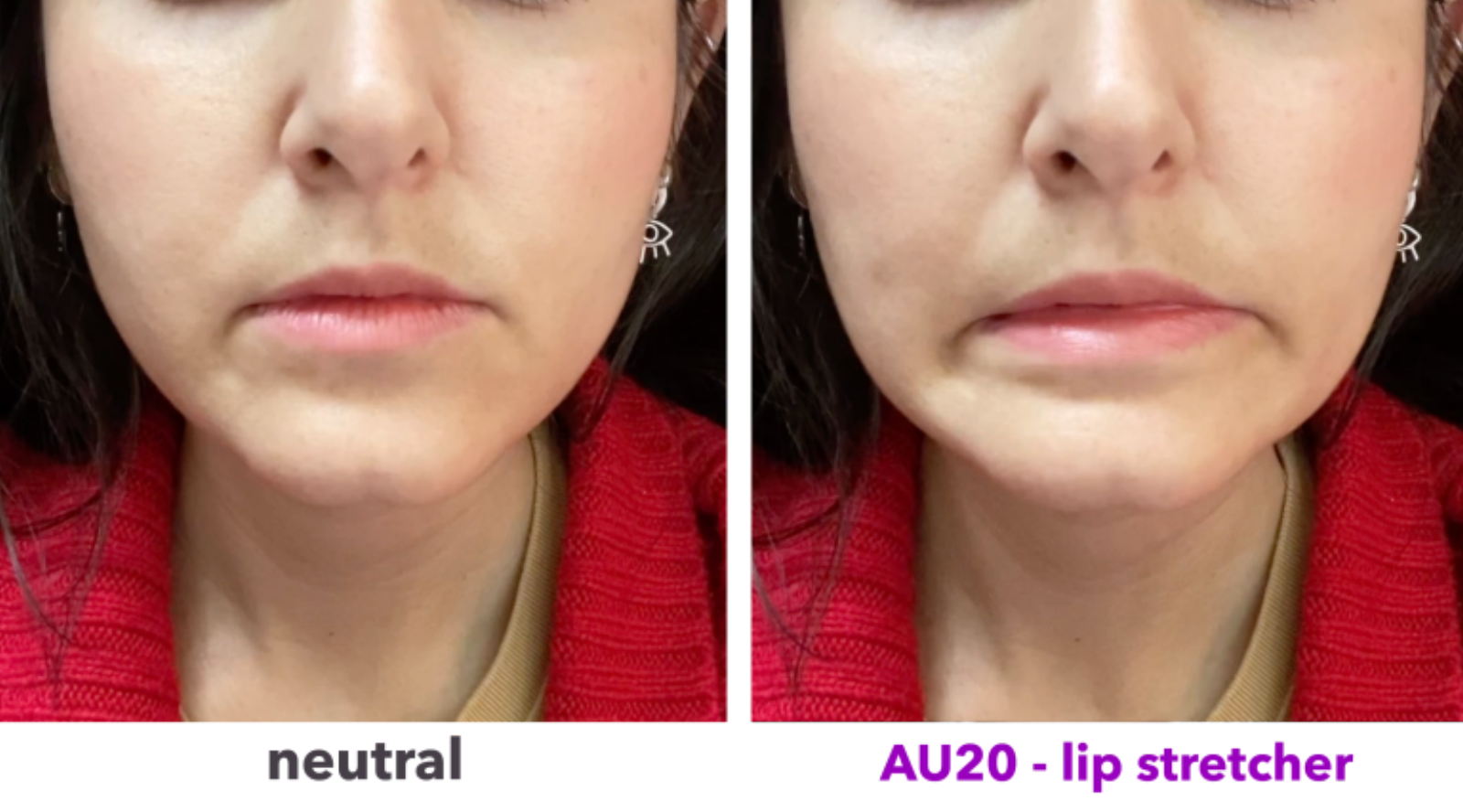}
  \end{subfigure}\\
  \begin{subfigure}{0.4\linewidth}
    \centering
    \includegraphics[width=\linewidth]{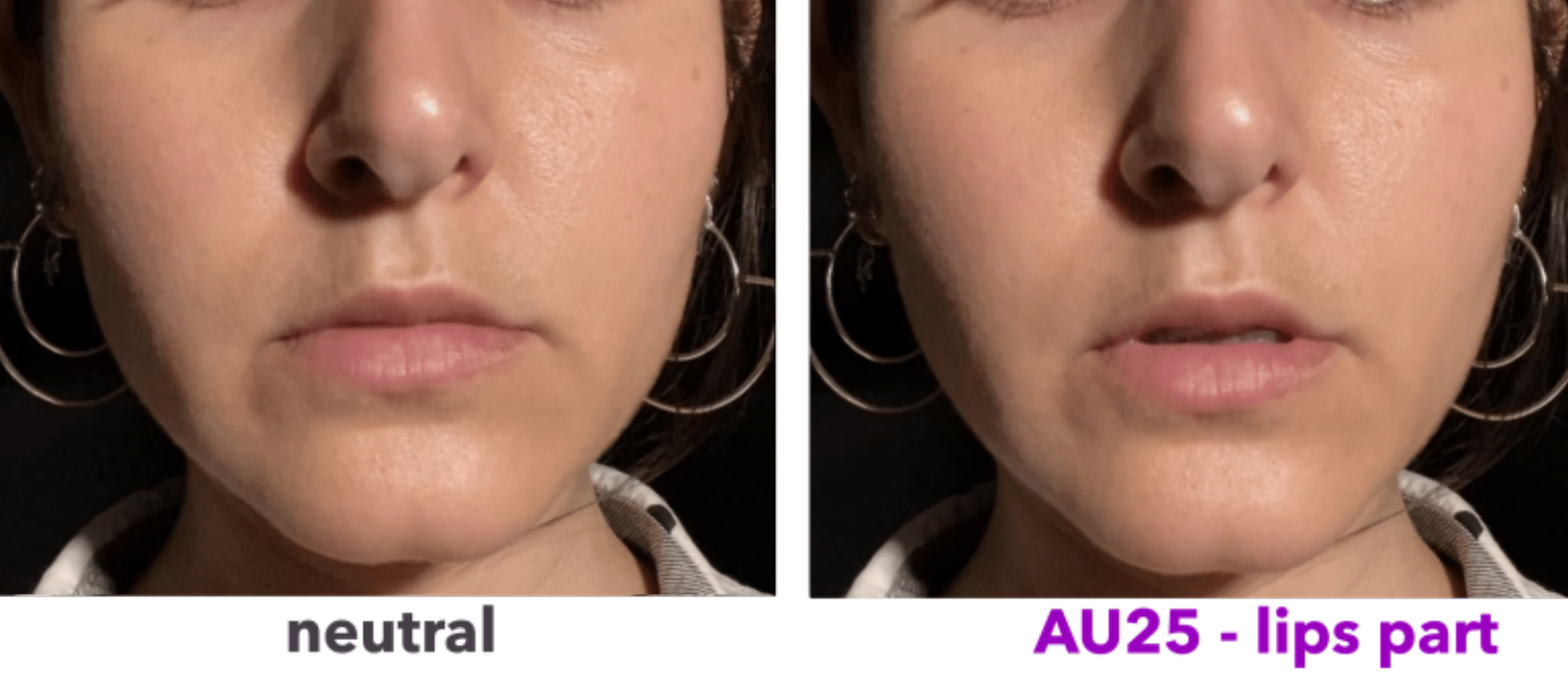}
  \end{subfigure}
  \hspace{0.08\linewidth}
  \begin{subfigure}{0.4\linewidth}
    \centering
    \includegraphics[width=\linewidth]{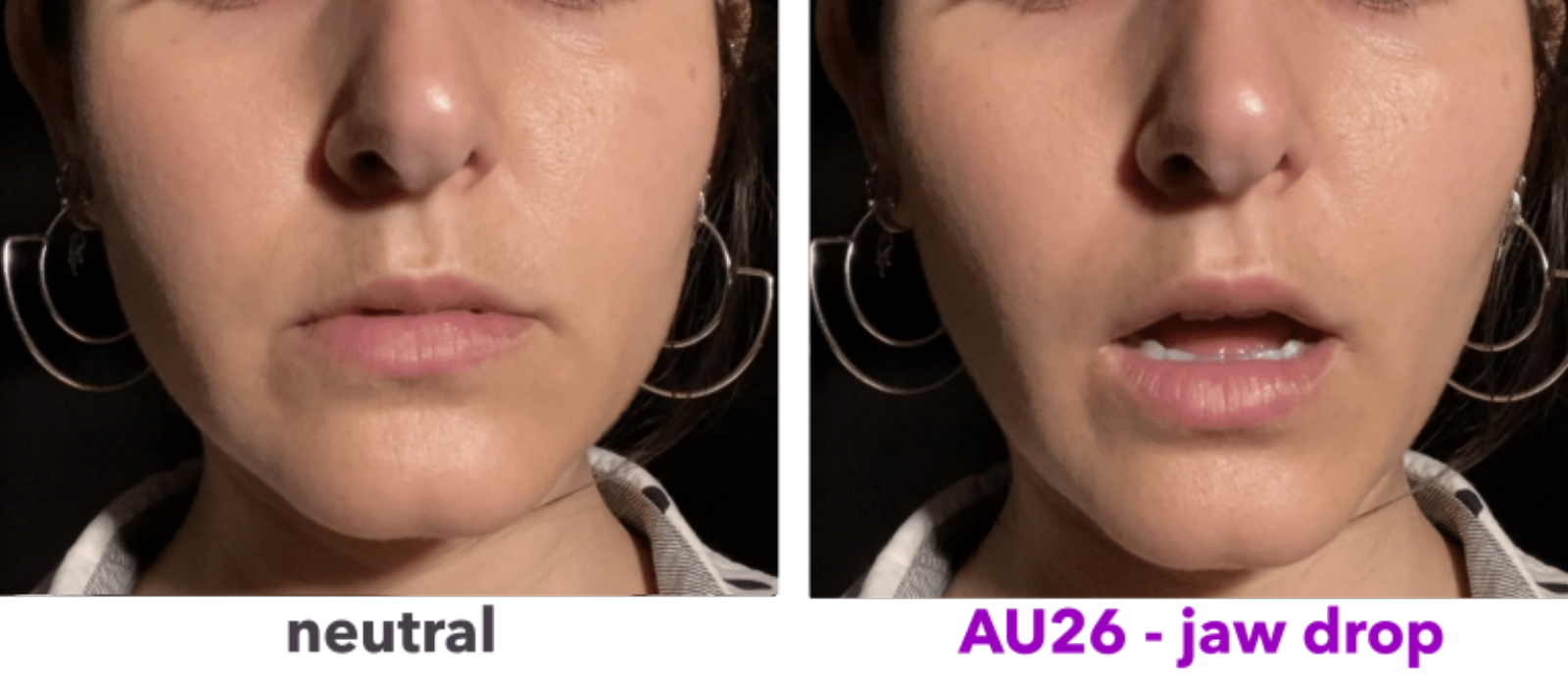}
  \end{subfigure}
  \caption{A visual reference guide for the analyzed AUs extracted from \cite{FACSlite}.}
  \label{fig:AUs}
\end{figure*}

The DISFA dataset \cite{mavadati2013disfa} contains facial video recordings of 27 participants' spontaneous facial expression while viewing video clips with approximately $130,000$ frames in total. Each frame is manually annotated by a human expert with intensities of AU1 (inner brow raiser), AU2 (outer brow raiser), AU4 (brow lowerer), AU5 (upper lid raiser), AU6 (cheek raiser), AU9 (nose wrinkler), AU12 (lip corner puller), AU15 (lip corner depressor), AU17 (chin raiser), AU20 (lip stretcher), AU25 (lips part), and AU26 (jaw drop) on a scale of 0 to 5. See \Cref{fig:AUs} for a visual reference guide for the analyzed AUs.

The DISFA+ dataset \cite{mavadati2016extended} is an extension of the DISFA dataset \cite{mavadati2013disfa}. It contains facial video recordings of 9 participants' posed and spontaneous facial expression with each frame being annotated with the same 12 AUs on a scale of 0 to 5.

\subsection{Video Preprocessing}

For each frame of the video, we first detect and crop the largest face in the image using MediaPipe \cite{lugaresi2019mediapipe} and then align the face with Pytorch Face Landmark Detection \cite{PFL}. Then, we further employ a combination of histogram equalization and linear mapping \cite{kuo2018compact} to increase the global contrast of the facial image. The facial images are resized to $112\times 112$ pixels before being fed into the neural network. Similar preprocessing is also applied to the videos we analyze in the FERGI dataset.

\subsection{Model Training}

We use the neural network IR-50 \cite{deng2019arcface} pre-trained on Glink360k \cite{an2022killing} and fine-tune it on the DISFA and DISFA+ datasets \cite{mavadati2013disfa,mavadati2016extended}. The last layer of the network is modified so that it outputs the estimation of the AUs in two formats: for estimating the intensity of the $i$th AU $y_{i}$, it outputs 1 value $\hat{y}_{i, \mathrm{reg}}$ representing the numerical estimation of the AU intensity (in the format of regression) and 5 values $\hat{y}_{i, \mathrm{class(1)}}$, $\hat{y}_{i, \mathrm{class(2)}}$, $\hat{y}_{i, \mathrm{class(3)}}$, $\hat{y}_{i, \mathrm{class(4)}}$, and $\hat{y}_{i, \mathrm{class(5)}}$ respectively representing the estimated probability of the AU intensity being higher than or equal to 1, 2, 3, 4, and 5 \cite{niu2016ordinal} (in the format of binary classifications). The loss function consists of three parts:
\begin{equation}
    E = E_\mathrm{reg,MSE} + E_\mathrm{reg,cos} + E_\mathrm{class},
\end{equation}
where $E_\mathrm{reg,MSE}$, $E_\mathrm{reg,cos}$, and $E_\mathrm{class}$ respectively represent a mean squared error (MSE) loss for the numerical estimations
\begin{equation}
    E_\mathrm{reg,MSE} = \Sigma_{i=1}^{n}w_{i,y_{i}}(y_{i}-\hat{y}_{i, \mathrm{reg}})^{2},
    \label{eqn:loss_reg_MSE}
\end{equation}
a cosine similarity loss for the numerical estimations
\begin{equation}
    E_\mathrm{reg,cos} = 1 - \frac{\Sigma_{i=1}^{n}y_{i}\hat{y}_{i, \mathrm{reg}}}{(\Sigma_{i=1}^{n}{y_{i}^{2}})(\Sigma_{i=1}^{n}\hat{y}_{i, \mathrm{reg}}^{2})},
    \label{eqn:loss_reg_cos}
\end{equation}
and a cross entropy loss for the binary classification estimations
\begin{equation}
    E_\mathrm{class} = \Sigma_{i=1}^{n}\Sigma_{j=1}^{5}w_{i,j,\chi_{y_{i}\geq j}}CE(\chi_{y_{i}\geq j}, \sigma(\hat{y}_{i, \mathrm{class(j)}})),
    \label{eqn:loss_class}
\end{equation}
with the cross entropy function being 
\begin{equation}
    CE(y,\hat{y})=-[y_i \log(\hat{y}_i) + (1 - y_i) \log(1 - \hat{y}_i)].
    \label{eqn:cross_entropy}
\end{equation}

The weights for the MSE loss and those for the cross entropy loss are both inverse-frequency weighted and normalized within each AU for addressing class imbalance in the datasets (substantially higher number of occurrences for low AU intensities).  However, since the number of occurrences of high intensities are too few for most AUs (resulting in too high weights for the MSE loss if used directly), we ``bin'' the intensities into 2 groups, and each group shares the same weight. Specifically, for the MSE loss, we apply one weight for occurrences of intensities of 0 and 1 and another weight for occurrences of intensities of 2, 3, 4, and 5, and these weights are computed based on the total number of occurrences within each intensity group.

Specifically, the weights for the MSE loss are defined as 
\begin{align}
  w_{i,j} = \begin{cases}
    \frac{2\cdot \frac{1}{\Sigma_{j'=0}^{1}n_{i,j'}}}{2\cdot\frac{1}{\Sigma_{j'=0}^{1}n_{i,j'}} + 4\cdot\frac{1}{\Sigma_{j'=2}^{5}n_{i,j'}}}, & \text{for $j=0,1$} \\
    \frac{4\cdot\frac{1}{\Sigma_{j'=2}^{5}n_{i,j'}}}{2\cdot\frac{1}{\Sigma_{j'=0}^{1}n_{i,j'}} + 4\cdot \frac{1}{\Sigma_{j'=2}^{5}n_{i,j'}}}, & \text{for $j=2,3,4,5$} \label{eqn:weights_reg}
  \end{cases}
\end{align}
while the weights for the cross entropy loss are defined as
\begin{align}
    \begin{cases}
    w_{i,j,1} = \frac{\frac{1}{\Sigma_{j'=j}^{5}n_{i,j'}}}{\Sigma_{j''=1}^{5}(\frac{1}{\Sigma_{j'=0}^{j''-1}n_{i,j'}} + \frac{1}{\Sigma_{j'=j''}^{5}n_{i,j'}})} \\
    w_{i,j,0} = \frac{\frac{1}{\Sigma_{j'=0}^{j-1}n_{i,j'}}}{\Sigma_{j''=1}^{5}(\frac{1}{\Sigma_{j'=0}^{j''-1}n_{i,j'}} + \frac{1}{\Sigma_{j'=j''}^{5}n_{i,j'}})},
    \label{eqn:weights_class}
    \end{cases}
\end{align}
where $n_{i,j}$ represents the number occurrences of the $i$th AU with an intensity of $j$.

Notably, although we train the neural network to learn both numerical estimations and binary classification estimations of the AU intensities, only the numerical estimations are used in model inference.

For model training, we employ the Adam optimizer with an initial learning rate of $10^{-4}$ for parameters of the last layer and an $10^{-5}$ for other parameters, a weight decay of $5\times 10^{-4}$, and a batch size of 64. We train the model on all data from the DISFA and DISFA+ datasets \cite{mavadati2013disfa,mavadati2016extended} for a total of 3 epochs using a single NVIDIA GeForce GTX 1080Ti 11G GPU for about 1 hour.
\section{Data Filtering}
This section provides the details of frame exclusion not provided in \Cref{sec:DataFiltering}.

We aim to exclude frames with occlusions and frames with off-angle pitch or yaw. To achieve the goal, we follow the following three exclusion rules:
\begin{itemize}
    \item Frames with a face detection confidence score (FDCS) lower than 0.9 are excluded. The FDCS is given by MediaPipe \cite{lugaresi2019mediapipe}. Low FDCS is likely caused by occlusions.
    
    \item Frames with a yaw indicative ratio (YIR) out of the range of $[0.3, 0.7]$. The YIR is computed as
    \begin{equation}
        \textrm{YIR} = \frac{d_\textrm{eye-edge,left}}{d_\textrm{eye-edge,left}+d_\textrm{eye-edge,right}},
        \label{eqn:YIR}
    \end{equation}
    where $d_\textrm{eye-edge,left}$ represents the horizontal distance between the left eye and the left edge of the face while $d_\textrm{eye-edge,right}$ represents the horizontal distance between the right eye and the right edge of the face, both of which are computed based on the facial landmarks detected with \cite{PFL}.
    
    \item Frames with a pitch indicative ratio (PIR) out of the range of $[0.55, 0.85]$. The PIR is computed as
    \begin{equation}
        \textrm{PIR} = \frac{d_\textrm{nostrils-eyes}}{d_\textrm{eyes}},
        \label{eqn:PIR}
    \end{equation}
    where $d_\textrm{nostrils-eyes}$ represents the vertical distance between the center of the nostrils and the center of the eyes
    while $d_\textrm{eyes}$ represents the horizontal distance between the two eyes, both of which are computed based on the facial landmarks detected with \cite{PFL}.
\end{itemize}
\section{Experiments}

\subsection{FAU-Net Analysis}

\begin{figure*}[tb]
  \centering
  \begin{subfigure}{0.24\linewidth}
    \includegraphics[width=\linewidth]{figures/node_1_weights.pdf}
    \caption{${\color{red} w_{h(1)-o}\approx -0.58}$}
    \label{fig:node_1_weights_supplementary}
  \end{subfigure}
  \centering
  \begin{subfigure}{0.24\linewidth}
    \includegraphics[width=\linewidth]{figures/node_2_weights.pdf}
    \caption{${\color{red} w_{h(2)-o}\approx -0.53}$}
    \label{fig:node_2_weights_supplementary}
  \end{subfigure}
  \centering
  \begin{subfigure}{0.24\linewidth}
    \includegraphics[width=\linewidth]{figures/node_3_weights.pdf}
    \caption{${\color{red} w_{h(3)-o}\approx -0.50}$}
    \label{fig:node_3_weights_supplementary}
  \end{subfigure}
  \centering
  \begin{subfigure}{0.24\linewidth}
    \includegraphics[width=\linewidth]{figures/node_4_weights.pdf}
    \caption{${\color{red} w_{h(4)-o}\approx -0.49}$}
    \label{fig:node_4_weights_supplementary}
  \end{subfigure}
  \centering
  \begin{subfigure}{0.24\linewidth}
    \includegraphics[width=\linewidth]{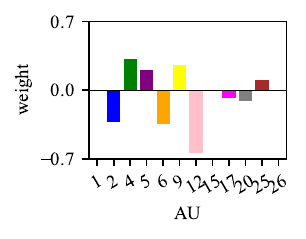}
    \caption{${\color{red} w_{h(5)-o}\approx -0.46}$}
    \label{fig:node_5_weights_supplementary}
  \end{subfigure}
  \centering
  \begin{subfigure}{0.24\linewidth}
    \includegraphics[width=\linewidth]{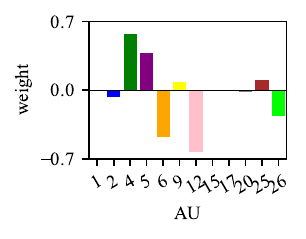}
    \caption{${\color{red} w_{h(6)-o}\approx -0.45}$}
    \label{fig:node_6_weights_supplementary}
  \end{subfigure}
  \centering
  \begin{subfigure}{0.24\linewidth}
    \includegraphics[width=\linewidth]{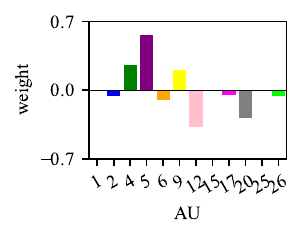}
    \caption{${\color{red} w_{h(7)-o}\approx -0.45}$}
    \label{fig:node_7_weights_supplementary}
  \end{subfigure}
  \centering
  \begin{subfigure}{0.24\linewidth}
    \includegraphics[width=\linewidth]{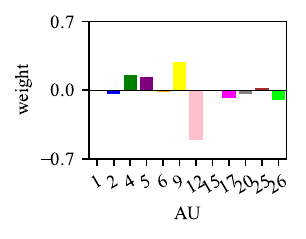}
    \caption{${\color{red} w_{h(8)-o}\approx -0.42}$}
    \label{fig:node_8_weights_supplementary}
  \end{subfigure}
  \centering
  \begin{subfigure}{0.24\linewidth}
    \includegraphics[width=\linewidth]{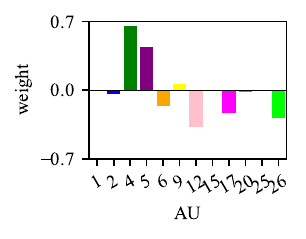}
    \caption{${\color{red} w_{h(9)-o}\approx -0.37}$}
    \label{fig:node_9_weights_supplementary}
  \end{subfigure}
  \centering
  \begin{subfigure}{0.24\linewidth}
    \includegraphics[width=\linewidth]{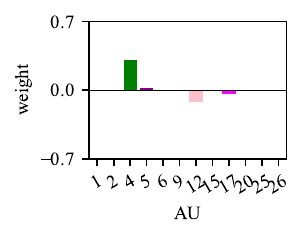}
    \caption{${\color{red} w_{h(10)-o}\approx -0.22}$}
    \label{fig:node_10_weights_supplementary}
  \end{subfigure}
  \centering
  \begin{subfigure}{0.24\linewidth}
    \includegraphics[width=\linewidth]{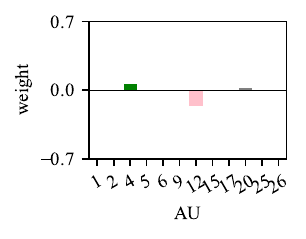}
    \caption{${\color{red} w_{h(11)-o}\approx -0.13}$}
    \label{fig:node_11_weights_supplementary}
  \end{subfigure}
  \centering
  \begin{subfigure}{0.24\linewidth}
    \includegraphics[width=\linewidth]{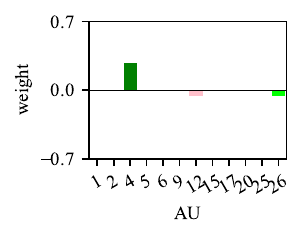}
    \caption{${\color{red} w_{h(12)-o}\approx -0.12}$}
    \label{fig:node_12_weights_supplementary}
  \end{subfigure}
  \centering
  \begin{subfigure}{0.24\linewidth}
    \includegraphics[width=\linewidth]{figures/node_13_weights.pdf}
    \caption{${\color{darkgreen} w_{h(13)-o}\approx 0.30}$}
    \label{fig:node_13_weights_supplementary}
  \end{subfigure}
  \centering
  \begin{subfigure}{0.24\linewidth}
    \includegraphics[width=\linewidth]{figures/node_14_weights.pdf}
    \caption{${\color{darkgreen} w_{h(14)-o}\approx 0.31}$}
    \label{fig:node_14_weights_supplementary}
  \end{subfigure}
  \centering
  \begin{subfigure}{0.24\linewidth}
    \includegraphics[width=\linewidth]{figures/node_15_weights.pdf}
    \caption{${\color{darkgreen} w_{h(15)-o}\approx 0.42}$}
    \label{fig:node_15_weights_supplementary}
  \end{subfigure}
  \centering
  \begin{subfigure}{0.24\linewidth}
    \includegraphics[width=\linewidth]{figures/node_16_weights.pdf}
    \caption{${\color{darkgreen} w_{h(16)-o}\approx 0.56}$}
    \label{fig:node_16_weights_supplementary}
  \end{subfigure}

  \caption{\textbf{Weights of hidden nodes.} Each subfigure shows the weights from the input preprocessed activation values of the 12 AUs to a hidden node in the FAU-Net, with each subcaption illustrating the weight from the hidden node to the output node (red/green for negative/positive weights). The hidden nodes are ordered by the weights from them to the output node in this figure.}
  \label{fig:weights_of_hidden_nodes_full}
\end{figure*}

A full version of \Cref{fig:weights_of_hidden_nodes} showing the weights of the hidden nodes in the FAU-Net is shown as \Cref{fig:weights_of_hidden_nodes_full}.

\end{document}